\def\year{2022}\relax
\documentclass[letterpaper]{article} 
\usepackage{aaai22}  
\usepackage{times}  
\usepackage{helvet}  
\usepackage{courier}  
\usepackage[hyphens]{url}  
\usepackage{graphicx} 
\usepackage{natbib}  
\usepackage{caption} 
\DeclareCaptionStyle{ruled}{labelfont=normalfont,labelsep=colon,strut=off} 
\usepackage{algorithm}

\usepackage{newfloat}
\usepackage{listings}
\floatstyle{ruled}
\newfloat{listing}{tb}{lst}{}
\floatname{listing}{Listing}
\usepackage{amsmath,amsfonts}
\usepackage{tabularx}
\usepackage{algpseudocode}%
\usepackage{multirow}
\usepackage{subfigure}
\usepackage{booktabs}       
\usepackage{nicefrac}       
\usepackage{microtype}      

\usepackage{tikz}
\usetikzlibrary{shapes,arrows}
\tikzstyle{decision} = [diamond, draw, fill=red!15, text width=4.5em, text badly centered, node distance=3cm, inner sep=0pt]
\tikzstyle{block} = [rectangle, draw, fill=blue!15, 
    text width=5em, text centered, rounded corners, minimum height=4em]
\tikzstyle{line} = [draw, -latex']
\tikzstyle{cloud} = [draw, ellipse,fill=red!20, node distance=3cm,
    minimum height=2em]

\title{The Impact of Reinitialization on Generalization\\ in Convolutional Neural Networks}
\author{
  Ibrahim~Alabdulmohsin, Hartmut~Maennel, Daniel~Keysers
}
\affiliations{
    Google Research\\ 8002 Z\"urich, Switzerland\\
    \texttt{\{ibomohsin,hartmutm,keysers\}@google.com}
}

\begin{document}

\newcommand{\BE}{{\mathbb E}}%
\newcommand{\BH}{{\mathbb H}}%
\newcommand{\BI}{{\mathbb I}}%
\newcommand{\BN}{{\mathbb N}}%
\newcommand{\BP}{{\mathbb P}}%
\newcommand{\BQ}{{\mathbb Q}}%
\newcommand{\BR}{{\mathbb R}}%
\newcommand{\BZ}{{\mathbb Z}}%

\newcommand{\calA}{{\mathcal A}}%
\newcommand{\calB}{{\mathcal B}}%
\newcommand{\calC}{{\mathcal C}}%
\newcommand{\calD}{{\mathcal D}}%
\newcommand{\calE}{{\mathcal E}}%
\newcommand{\calF}{{\mathcal F}}%
\newcommand{\calG}{{\mathcal G}}%
\newcommand{\calH}{{\mathcal H}}%
\newcommand{\calI}{{\mathcal I}}%
\newcommand{\calJ}{{\mathcal J}}%
\newcommand{\calL}{{\mathcal L}}%
\newcommand{\calM}{{\mathcal M}}%
\newcommand{\calN}{{\mathcal N}}%
\newcommand{\calP}{{\mathcal P}}%
\newcommand{\calS}{{\mathcal S}}%
\newcommand{\calT}{{\mathcal T}}%
\newcommand{\calV}{{\mathcal V}}%
\newcommand{\calX}{{\mathcal X}}%
\newcommand{\calY}{{\mathcal Y}}%
\newcommand{\calZ}{{\mathcal Z}}%

\newcommand{\xb}{\textbf{x}}%
\newcommand{\yb}{\textbf{y}}%
\newcommand{\zb}{\textbf{z}}%
\newcommand{\wb}{\textbf{w}}%

\newcommand{\reinit}{{\sc lw }}
\newcommand{\reinitns}{{\sc lw}}

\newcommand{\dsd}{{\sc dsd }}
\newcommand{\dsdns}{{\sc dsd}}

\newcommand{\wels}{{\sc wels }}
\newcommand{\welsns}{{\sc wels}}

\newcommand{\welsr}{{\sc welsr }}
\newcommand{\welsrns}{{\sc welsr}}

\newcommand{\fc}{{\sc fc }}
\newcommand{\fcns}{{\sc fc}}

\maketitle

\begin{abstract}
Recent results suggest that reinitializing a subset of the parameters of a neural network during training can improve generalization, particularly for small training sets. We study the impact of different reinitialization methods in several convolutional architectures across 12 benchmark image classification datasets, analyzing their potential gains and highlighting limitations. We also introduce a new layerwise reinitialization algorithm that outperforms previous methods and suggest explanations of the observed improved generalization. First, we show that layerwise reinitialization increases the margin on the training examples without increasing the norm of the weights, hence leading to an improvement in margin-based generalization bounds for neural networks. Second, we demonstrate that it settles in flatter local minima of the loss surface. Third, it encourages learning general rules and discourages memorization by placing emphasis on the lower layers of the neural network. Our takeaway message is that the accuracy of convolutional neural networks can be improved for small datasets using bottom-up layerwise reinitialization, where the number of reinitialized layers may vary depending on the available compute budget. 
\end{abstract}

\section{Introduction}
\label{sect::intro}
Deep neural networks have demonstrated state-of-the-art performance over many classification tasks. While often highly overparameterized, modern deep neural network architectures exhibit a remarkable ability to generalize beyond the training sample even when trained without any explicit form of regularization \citep{Zhang2017}. A large body of work has been devoted to offering insights into this ``benign'' overfitting phenomenon, including explanations based on the margin \citep{bartlett1998sample,neyshabur2015norm,bartlett2017spectrally,neyshabur2017exploring,arora2018stronger,soudry2018implicit}, the curvature of the local minima \citep{keskar2016large,chaudhari2019entropy,neyshabur2020being}, and the speed of convergence \citep{hardt2016train}, among others. 

Recently, however, a number of works suggest that generalization in convolutional neural networks (CNNs) could be improved further 
using reinitialization. Precisely, let $\textbf{w}\in\mathbb{R}^d$ be a vector that contains all of the parameters in a neural network (e.g.\ filters in convolutional layers and weight matrices in fully-connected layers). Let $\textbf{s}\in\{0,\,1\}^d$ be a binary mask that is generated at random according to some probability mass function. Then, ``reinitialization'' refers to the practice of selecting a subset of the parameters and reinitializing those during training:
\begin{equation}\label{reinit_eq}
    \textbf{w} \;\leftarrow\; (1-\textbf{s}) \odot \textbf{w} \,+\, \textbf{s}\odot \eta,
\end{equation}
where $\odot$ is an element-wise multiplication and $\eta$ is a random initialization of the model parameters. In the following, we refer to the update in (\ref{reinit_eq}) as a  ``reinitialization round.'' Various reinitialization methods differ in how the binary mask $\textbf{s}$ is selected. Four prototypical approaches are:
\begin{itemize}
    \item \textbf{Random subset}: A random subset of the parameters of a fixed size is chosen uniformly at random in each round. This includes, for example, the random weight level splitting (\welsrns) method studied in \cite{taha2021knowledge}, in which about 20\% of the parameters are selected for reinitialization.

    \item \textbf{Weight magnitudes}: The smallest parameters in terms of their absolute magnitudes are reinitialized at each round. This can be interpreted as a generalization to the sparse-dense-sparse (\dsdns)  workflow of \cite{han2016dsd} in which reinitialization occurs only once.
    
    \item \textbf{Fixed subset}: A subset is chosen at random initially prior to training and is fixed afterwards. This corresponds to the weight level splitting (\welsns) method of \cite{taha2021knowledge}.
    
    \item \textbf{Fully-connected layers}: Only the last fully-connected layers are reinitialized. This includes, for example, the method proposed in \cite{li2020rifle}. In \cite{zhao2018retraining}, only the classifier head is reinitialized.
\end{itemize}
\noindent We denote these four methods as \welsrns, \dsdns, \welsns, and \fcns, respectively. In addition, we denote the baseline method of training once until convergence as {\sc bl}. 

In this paper, we introduce a new reinitialization algorithm, which we denote as \reinit for its LayerWise approach. The new algorithm is motivated by the common observation that lower layers in the neural network tend to learn general rules while upper layers specialize \citep{yosinski2014transferable,arpit2017closer,raghu2019transfusion,maennel2020neural,baldock2021deep}. While all reinitialization methods improve generalization in CNNs, we demonstrate in Section~\ref{sect::experiments} that \reinit often outperforms the other methods. It encourages learning general rules by placing more emphasis on training the early layers of the neural network. 

\begin{figure*}[tb]
\centering
\includegraphics[width=1.98\columnwidth]{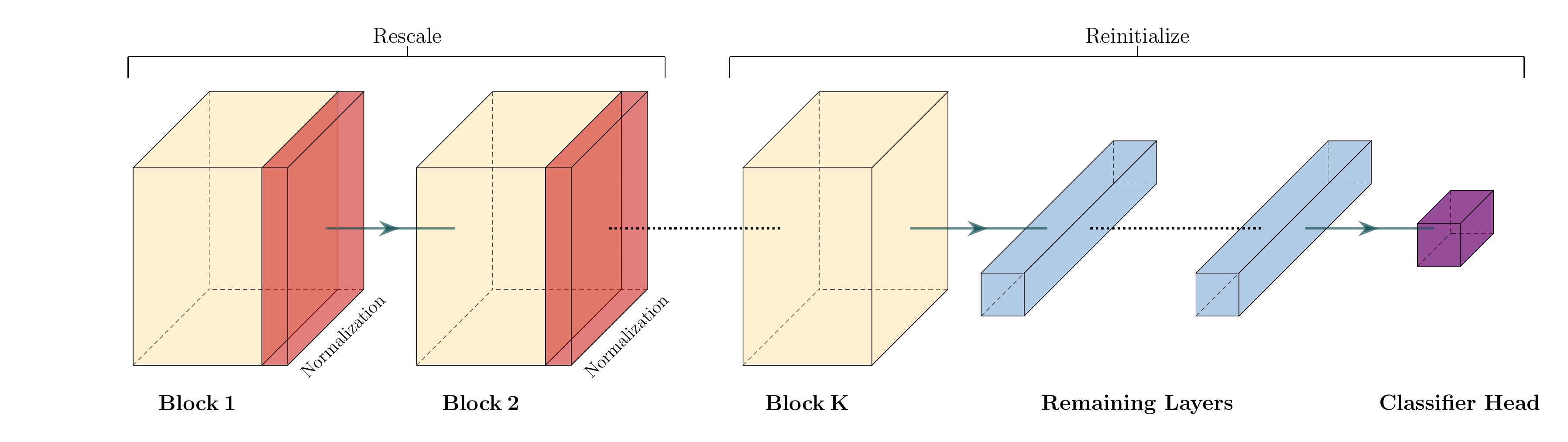} \vspace{-3ex}
\caption{Given a deep neural network starting with $K$ convolutional blocks followed by other layers, \reinit proceeds sequentially from bottom to top (see Algorithm~\ref{algorithm}).  When in block $k$ (e.g. $k=2$ in the figure above), the weights of all early blocks $\{1,\ldots, k\}$ are rescaled while subsequent layers are reinitialized. In addition, a normalization layer is inserted following block $K$.}
\label{fig:reinit}
\end{figure*}

\begin{itemize}
    \item \textbf{Layerwise}: A convolutional neural network is partitioned into $K$ blocks (see Figure~\ref{fig:reinit} and Algorithm~\ref{algorithm}). At round $k$, the parameters at the lowest $K$ blocks are rescaled back to their original norm during initialization while the rest of the network is reinitialized. In addition, a new normalization layer is inserted/updated following block $K$. This is repeated for a total of $N\ge 1$ iterations for each block.
\end{itemize}

\begin{algorithm}[tb]
 \small
 \vspace{1mm}
 \textbf{Input:} (1) Neural network with identified sequence of $K\ge1$ conv blocks; (2) Training dataset; (3) $N\ge 1$.
 \vspace{1mm}
 
 \textbf{Output:} Trained model parameters.
 \vspace{1mm}
 
 \textbf{Training:}
 \vspace{1mm}
    \begin{algorithmic}[1]
     \State Initialize the neural network architecture and record the scales of each layer during initialization;
     \For{$k\in(1,2,\ldots,K)$}
        \For{$n\in(1,2,\ldots,N)$}
            \State Rescale the weights of all blocks $\{1,2,\ldots,k\}$;
            \State Compute $Z$: the output of Block $k$ of a sample $X$;
            \State Compute $\mu, \sigma\in\mathbb{R}$: mean and standard deviation of $Z$;
            \If{$n=1$}
            \State Insert a lambda layer $\lambda x: (x-\mu)/\sigma$ after block $k$;
            \Else
            \State Update lambda layer with new values of $\mu$ and $\sigma$;
            \EndIf
            \State Reinitialize all layers above block $k$;
            \State Fine-tune the entire model until convergence;
        \EndFor
     \EndFor
    \end{algorithmic}
\caption{Pseudocode of \reinit}\label{algorithm}
\end{algorithm}
 
 It is worth noting that \fc is a special case of \reinitns, in which $K=1$ and $N>1$. Besides the prominent role of reinitialization, \reinit includes normalization and rescaling, which we show in an ablation study in Appendix~\ref{appendix:ablation} to be important. Next, we illustrate the basic principles of these  reinitialization methods on a minimal example with synthetic data.

\subsection{Synthetic Data Example}\label{sect::syntheticdata}

\begin{table}[tbp]
    \centering
    \footnotesize
    \caption{Test accuracy [\%] for the synthetic data experiment of Section~\ref{sect::syntheticdata} with different signal strengths $\alpha$ and different reinitialization methods. We observe that all initialization methods (with the exception of \dsdns) improve generalization in this example setting with \reinit performing best. In addition, reinitialization methods also tend to reduce the variance of the test accuracy.}
    \label{tab:synthetic_data_table}
    \begin{tabularx}{\columnwidth}{@{}X|XXXXXX@{\hspace{0.5ex}}}
    \toprule
      \em $\alpha$ & \em {\sc bl} & \em \welsr & \em \dsd & \em \wels &\em \fc &\em \reinit\rule{1em}{0pt}\\\midrule
      0.5 &$20.3$ &$24.6$ &$22.9$ &$23.1$ &$23.6$ &$25.2$\\
      &($\pm 0.6$) &($\pm 1.0$) &($\pm 0.6$) &($\pm 1.4$) &($\pm 3.0$) &($\pm 0.8$)\\[0.6em]
      $1.0$ &$50.7$ &$72.9$ &$53.4$ &$66.1$ &$68.6$ &$72.3$ \\
      &($\pm 5.4$) &($\pm 0.9$) &($\pm 0.7$) &($\pm 2.1$) &($\pm 2.1$) &($\pm 3.6$) \\[0.6em]
      $2.0$ &$94.6$ &$98.2$ &$90.3$ &$96.8$ &$99.0$ &$99.8$   \\
      &($\pm 2.0$) &($\pm 0.4$) &($\pm 1.4$) &($\pm 0.1$) &($\pm 0.2$) &($\pm 0.2$)  
      \\ \bottomrule
\end{tabularx}
\end{table}

\paragraph{Setup.}Let $\xb\in\BR^{128}$ be the instance and $\yb$ be its label, which is sampled uniformly at random from the set $\{0,1,...,7\}$. For the instances, on the other hand, each of the first 3 coordinates of $\xb$ is chosen from $\{-1,1\}$ to encode the label~$\yb$ appropriately in binary form. For example, instances that belong to the class 0 would have their first three coordinates as $(-1, -1, -1)$, whereas instances in class 5 would have $(1, -1, 1)$. Consequently, the first three coordinates of an instance correspond to its ``signal.'' The remaining 125 entries of $\xb$ are randomly sampled i.i.d. from $\calN(0,\,1)$.

Although we focus in this work on convolutional neural networks (CNN), we use a multilayer perceptron (MLP) in this synthetic data experiment for illustration purposes because the inputs are not images but generic feature vectors. The MLP contains two hidden layers of 32 neurons with ReLU activations \citep{nair2010rectified} followed by a classifier head with softmax activations. It optimizes the cross--entropy loss.
We train on 256 examples using  gradient descent  with a learning rate 0.05.

\paragraph{Methods.}Treating every layer as a block, we have $K=3$. If $200$ training steps are used per round of reinitialization and $N=3$, \reinit trains the model once for $200$ steps after which the 2\textsuperscript{nd} and 3\textsuperscript{rd} layers are reinitialized (in addition to rescaling and normalization).  This is carried out $N=3$ times in the first layer before $k$ is incremented. The same process is repeated on each layer making a total of $200\times N\times K=1,800$ training steps overall. In \welsns, \welsrns, \dsdns, and \fcns, the model is trained for  $200$ steps before reinitialization is applied, and this is repeated $K\times N$ times for the same total of $1,800$ steps. The baseline method corresponds to training the model once without reinitialization for a total of $1,800$ training steps. 

\paragraph{Results.}When trained for 1,800 steps, the baseline ({\sc bl}) achieves  100\% training accuracy, but only around 51\% test accuracy. The large gap between training and test accuracy for such a simple task is reminiscent of the classical phenomenon of overfitting. Note that the number of training examples is 256, which is generally small for 128 features of equal variance. On the other hand, reinitialization improves accuracy as shown in Table \ref{tab:synthetic_data_table} even though these reinitialization methods do \emph{not} have access to any additional data and use the same optimizer and hyper-parameters as baseline training. The training accuracy is 100\% in all cases. We also observe that reinitialization tends to reduce the variance of the test accuracy (with respect to the random seed).

In the above experiment, both the signal part (first three coordinates) and the noise part (remaining coordinates) have the same scale (standard deviation 1). We can make the classification problem easier or harder by multiplying the signal part by a signal strength $\alpha>1$ or $\alpha<1$, respectively. We present the average test accuracy in Table~\ref{tab:synthetic_data_table} for a selection of values of $\alpha$ with $N=3$.  
Appendix~\ref{sect::appendix_toy_example} contains additional results when weight decay is added.

\subsection{Contribution}
In this work, we introduce a new layerwise reinitialization algorithm \reinitns, which outperforms previous methods. We provide two explanations, supported by experiments, for why it improves generalization in convolutional neural networks. First, we show that \reinit improves the margin on the training examples without increasing the norm of the weights, hence leading to an improvement in known margin-based generalization bounds in neural networks. Second, we show that \reinit settles in flatter local minima of the loss surface.

Furthermore, we provide a comprehensive study comparing previous reinitialization methods: First, we evaluate different methods within the same context. For example, the comparison in \cite{taha2021knowledge} uses only a single reinitialization round of the dense-sparse-dense approach (\dsdns), while \dsd can be extended to multiple rounds. Also, \cite{zhao2018retraining} uses an ensemble of classifiers when reinitializing the fully-connected layers, which could (at least partially) explain the improvement in performance. By contrast, we follow a coherent training protocol for all methods. Second, we use our empirical evaluation to analyze the effect of the experiment's design, such as augmentation, dropout, learning rate, and momentum. The goal is to determine if the effect of reinitialization could be achieved by tuning such settings. Third, we employ decision tree classifiers to identify when each reinitialization method is likely to outperform others. In summary, we:
\begin{enumerate}
    \item Introduce a new reinitialization method that is motivated by common observations of generalization and memorization effects across the neural network's layers. We show that it outperforms other methods with a statistically significant evidence at the 95\% level.
    \item Suggest two explanations, supported by experiments, for why \reinit is more successful at improving generalization in CNNs compared to other methods.
    \item Present a comprehensive evaluation study of reinitialization methods covering more than 2,000 experiments for four convolutional architectures: (1) simplified CNN, (2) VGG16 \citep{simonyan2014very}, (3) MobileNet \citep{howard2017mobilenets} and (4) ResNet50 \citep{He2016}. We conduct the evaluation over 12 benchmark image classification datasets (cf. Appendix~\ref{appendix::exp_setup}).
\end{enumerate}

\section{Related Work}\label{sect::related_work}
\paragraph{Reinitialization.}As stated earlier, a number of works suggest that reinitializing a subset of the neural network parameters during training can improve  generalization. This includes,
the dense-sparse-dense (\dsdns) training workflow  proposed by ~\cite{han2016dsd}, in which  reinitialization occurs only once during training. However, as the authors argue, the improvement in accuracy in \dsd could be attributed to the effect of introducing sparsity, not reinitialization. Another example is ``Knowledge Evolution'', including weight level splitting (\welsns) and its randomized version (\welsrns) \citep{taha2021knowledge}. It was noted that \wels outperformed \welsrns, which agrees with our observations. Finally, some recent works propose to reinitialize the fully-connected layers only~\citep{li2020rifle,zhao2018retraining}. In particular, reinitializing the last layer several times and combining the models into an ensemble can improve performance~\citep{zhao2018retraining}. However, the improvement in accuracy could (at least partially) be attributed to the ensemble of predictors,  not to reinitialization \emph{per se}. For fair comparison, we extend \dsd to multiple rounds of reinitialization and do not use an ensemble of predictors.

\paragraph{Generalization Bounds.}
Several generalization bounds for neural networks have been proposed in the literature. Of those, a prototypical approach is to bound the generalization gap by a particular measure of the \emph{size of weights} normalized by the \emph{margin} on the training set. Examples of measures of the size of weights include the product of the $\ell_1$ norms~\citep{bartlett1998sample} and the product of the Frobenius norms of layers~\citep{neyshabur2015norm}, among others~\citep{bartlett2017spectrally,neyshabur2017exploring,arora2018stronger}.  While such generalization bounds are often loose, they were found to be useful for ranking models~\citep{neyshabur2017exploring}. The fact that rich hypothesis spaces could still generalize if they yield a large margin over the training set was used  previously to explain the  performance of boosting~\citep{adaboost_margin}. In Section~\ref{sect::analysis}, we show that \reinit  boosts the margin on the training examples without increasing the size of the weights.

\paragraph{Flatness of the Local Minimum.}
Another important line of work examines the connection between generalization and the curvature of the loss at the local minimum \citep{keskar2016large,neyshabur2017exploring,sam2021}. Deep neural networks are known to converge to local minima with sparse eigenvalues ($>$94\% zeros) in their Hessian~\citep{chaudhari2019entropy}.
Informally, a flat local minimum is robust to data perturbation, and this robustness can, in turn, be connected to regularization~\citep{bishop1995training}. In fact, some of the benefits of transfer learning  were attributed to the flatness of the local minima~\citep{neyshabur2020being}. For a precise treatment, one may use the PAC-Bayes framework to derive a generalization bound that comprises of two terms: (1)~sharpness of the local minimum, and (2)~the weight norm over noise ratio~\citep{neyshabur2017exploring}. Similar terms also surface in the notion of ``local entropy''~\citep{chaudhari2019entropy}. We show in Section~\ref{sect::analysis} that \reinit improves both  terms.

\paragraph{Generalization vs. Memorization.}
Several works point out that early layers in a neural network tend to learn general-purpose representations whereas later layers specialize, e.g.~\citep{raghu2019transfusion,arpit2017closer, yosinski2014transferable,maennel2020neural}. This can be observed, for instance, using probes, in which classifiers are trained on the layer embeddings. As demonstrated in~\cite{cohen2018dnn} and \cite{baldock2021deep}, deep neural networks learn to separate classes at the early layers with real labels (generalization) but they only separate classes at later layers when the labels are random (memorization). 
One explanation for why  \reinit improves generalization is that it encourages learning general rules at early layers and discourages memorization at later layers.

\newcommand{\ts}{\rule{0pt}{1.9ex}}

\begin{table}[tp]
    \centering
    \footnotesize
    \caption{
    Test accuracy results [\%] for the five reinitialization methods across 12 benchmark datasets. Reinitialization improve generalization, in general, with \reinit often outperforming others. 
    These numbers include data augmentation; see Appendix~\ref{appendix::detailed_figures} for more results with different experiment settings. The abbreviations {\sc b,r,d,w,f,l} stand for baseline, \welsrns, \dsdns, \welsns, \fcns, and \reinitns, respectively. In \reinit, $N=1$. Also, every reinitialization method uses the same number of rounds $K$, where $K$ depends on the model architecture (cf.\ Appendix~\ref{appendix::exp_setup}). Note that we do not fine-tune the hyper-parameters $N$ and $K$. In the baseline method, which trains only once, we increase the maximum number of training steps to match that of reinitialization methods. 
    In \welsns, \welsrns, and \dsdns, we follow \cite{taha2021knowledge} in having 20\% of the parameters reinitialized. 
    }
    \label{tab:reinit_results_with_aug}
    \vspace{-1ex}
    \renewcommand{\arraystretch}{0.9}
    \begin{tabularx}{\columnwidth}{@{}llXXXXXX@{\hspace{0.95em}}}
    \toprule
      \em Dataset& \em Model\hspace{3em} & \em \sc b & \em \sc r & \em \sc d & \em \sc w &\em \sc f &\em \sc l\\\midrule
\multirow{4}{*}{\sc oxford-iiit}&\texttt{scnn} &15.1 &14.6 &14.3 &14.7 &16.0 &\bf16.6 \\&\texttt{vgg16} &22.0 &20.0 &20.9 &20.7 &\bf39.1 &29.8 \\&\texttt{mobilenet} &27.7 &24.1 &23.2 &24.3 &23.5 &\bf41.7 \\&\texttt{resnet50} &29.7 &33.3 &\bf39.1 &31.9 &36.6 &34.4 \\\hline
\multirow{4}{*}{\sc dogs}&\texttt{scnn} &\phantom{0}7.4 &\phantom{0}8.2 &\phantom{0}7.7 &\phantom{0}8.0 &\phantom{0}8.1 &\bf\phantom{0}8.6\ts \\&\texttt{vgg16} &16.0 &16.9 &15.5 &16.7 &\bf37.3 &31.2 \\&\texttt{mobilenet} &19.0 &19.5 &27.1 &18.8 &20.7 &\bf35.8 \\&\texttt{resnet50} &26.4 &30.7 &28.4 &35.2 &33.3 &\bf36.9 \\\hline
\multirow{4}{*}{\sc fmnist}&\texttt{scnn} &92.3 &92.3 &92.2 &92.2 &92.5 &\bf92.7\ts \\&\texttt{vgg16} &90.7 &92.3 &91.5 &92.3 &92.1 &\bf92.8 \\&\texttt{mobilenet} &91.7 &92.0 &92.4 &92.0 &91.5 &\bf92.7 \\&\texttt{resnet50} &92.9 &92.7 &93.0 &92.9 &93.2 &\bf93.4 \\\hline
\multirow{4}{*}{\sc cars196}&\texttt{scnn} &\phantom{0}5.6 &\phantom{0}6.0 &\phantom{0}5.3 &\phantom{0}5.5 &\bf\phantom{0}7.3 &\phantom{0}5.8\ts \\&\texttt{vgg16} &11.7 &10.7 &11.2 &11.4 &\bf43.6 &22.4 \\&\texttt{mobilenet} &\phantom{0}6.9 &16.2 &\phantom{0}9.1 &11.9 &13.8 &\bf44.0 \\&\texttt{resnet50} &21.8 &42.5 &33.2 &43.1 &36.7 &\bf43.6  \\\hline
\multirow{4}{*}{\sc mnist-cor}&\texttt{scnn} &\bf99.0 &\bf99.0 &\bf99.0 &\bf99.0 &\bf99.0 &\bf99.0\ts \\&\texttt{vgg16} &\bf99.0 &98.9 &98.9 &98.9 &\bf99.0 &\bf99.0 \\&\texttt{mobilenet} & 98.9 &98.8 &98.9 &98.8 &98.9 &\bf99.0 \\&\texttt{resnet50} &\bf99.1 &99.0 &\bf99.1 &\bf99.1 &99.0 &\bf99.1 \\\hline
\multirow{4}{*}{\sc cifar10-cor}&\texttt{scnn} &82.4 &83.8 &82.7 &83.3 &84.2 &\bf84.8\ts \\&\texttt{vgg16} &83.6 &86.5 &84.7 &85.8 &84.7 &\bf88.3 \\&\texttt{mobilenet} &85.8 &86.6 &\bf87.4 &\bf87.4 &86.0 &\bf87.4 \\&\texttt{resnet50} &88.0 &87.4 &\bf89.2 &89.1 &88.7 &88.3 \\\hline
\multirow{4}{*}{\sc cifar10}&\texttt{scnn} &82.7 &84.2 &82.5 &83.7 &\bf84.9 &84.3\ts \\&\texttt{vgg16} &84.7 &86.8 &84.9 &85.9 &85.0 &\bf88.9 \\&\texttt{mobilenet} &86.4 &86.6 &\bf88.1 &86.5 &86.3 &87.6 \\&\texttt{resnet50} &87.4 &87.5 &88.9 &\bf89.6 &89.2 &89.1 \\\hline\multirow{4}{*}{\sc caltech101}&\texttt{scnn} &50.1 &52.2 &52.2 &51.5 &\bf54.4 &52.8\ts \\&\texttt{vgg16} &55.9 &55.8 &57.1 &56.3 &\bf67.1 &59.1 \\&\texttt{mobilenet} &41.0 &41.5 &47.4 &41.8 &\bf48.0 &46.3 \\&\texttt{resnet50} &50.2 &51.9 &53.1 &\bf57.5 &52.1 &50.5 \\\hline\multirow{4}{*}{\sc cassava}&\texttt{scnn} &58.9 &60.6 &59.4 &62.2 &67.0 &\bf69.3\ts \\&\texttt{vgg16} &70.3 &70.0 &70.0 &71.2 &\bf71.5 &68.3 \\&\texttt{mobilenet} &62.3 &72.8 &80.1 &76.1 &77.3 &\bf81.1 \\&\texttt{resnet50} &46.5 &79.2 &\bf82.9 &82.6 &77.6 &73.9 \\\hline\multirow{4}{*}{\sc cmaterdb}&\texttt{scnn} &96.2 &97.0 &96.6 &97.3 &\bf97.4 &97.1\ts \\&\texttt{vgg16} &97.5 &\bf98.2 &97.1 &97.5 &97.5 &97.5 \\&\texttt{mobilenet} &97.9 &97.4 &97.7 &\bf98.0 &97.5 &97.0 \\&\texttt{resnet50} &97.4 &97.9 &97.7 &97.4 &97.7 &\bf98.4 \\\hline\multirow{4}{*}{\sc birds2010}&\texttt{scnn} &\phantom{0}3.8 &\phantom{0}3.7 &\phantom{0}4.0 &\phantom{0}3.2 &\bf\phantom{0}4.2 &\phantom{0}3.8\ts \\&\texttt{vgg16} &\phantom{0}5.5 &\phantom{0}6.5 &\phantom{0}5.6 &\phantom{0}5.8 &\bf13.2 &\phantom{0}9.8 \\&\texttt{mobilenet} &\phantom{0}6.6 &\phantom{0}9.0 &\phantom{0}8.1 &\phantom{0}6.2 &\phantom{0}7.4 &\bf\phantom{0}9.8 \\&\texttt{resnet50} &\phantom{0}8.5 &12.5 &11.2 &\bf13.0 &11.9 &10.6 \\\hline\multirow{4}{*}{\sc cifar100}&\texttt{scnn} &53.9 &55.0 &53.2 &53.8 &\bf58.0 &\bf58.0\ts \\&\texttt{vgg16} &53.2 &60.2 &53.1 &57.6 &55.2 &\bf64.3 \\&\texttt{mobilenet} &57.3 &\bf59.3 &60.6 &65.2 &57.1 &57.8 \\&\texttt{resnet50} &59.9 &60.0 &61.8 &\bf62.1 &60.3 &60.7 \\\bottomrule
    \end{tabularx}
\end{table}

\section{Empirical Study}\label{sect::experiments}
We begin by evaluating the performance of the five reinitialization methods discussed in Section~\ref{sect::intro} for four convolutional architectures on 12 benchmark datasets including, for example, CIFAR10/100~\citep{Krizhevsky09learningmultiple}, Fashion MNIST \citep{DBLP:journals/corr/abs-1708-07747}, and Caltech101 \citep{FeiFei2004LearningGV}, among others (see Appendix~\ref{appendix::exp_setup} for details). All images are resized to $224\times 224$. 
The architectures are (1)~simplified CNN, (2)~VGG16 \citep{simonyan2014very}, (3)~MobileNet \citep{howard2017mobilenets} and (4)~ResNet50 \citep{He2016}. We denote these by \texttt{scnn}, \texttt{vgg16}, \texttt{ mobilenet}, and \texttt{resnet50}, respectively. 
We use He-initialization~\citep{he2015delving} for all models unless stated otherwise. 

To recall, every reinitialization method trains the same model on the same dataset for several rounds. After each round, a binary mask of the model parameters is selected according to the reinitialization criteria and the update in Eq.~(\ref{reinit_eq}) is applied for some random initialization $\eta$. After that, the model is fine-tuned on the same data. Blocks \reinitns correspond to the standard blocks of the architecture (e.g. a block in Figure \ref{fig:reinit} would correspond to either an identity or a convolutional block in ResNet50). Also, 10\% of the training split is reserved as validation set, which is used for early stopping in all methods.

To evaluate the relative performance of the reinitialization methods
we perform a set of experiments in which we \textit{fix} the hyperparameters for all 
architectures and datasets to the same values. The hyperparameters were chosen to
work reasonably well across all combinations; in particular they enable reaching 100\% training accuracy in all cases.
We use SGD with an initial learning rate of 0.003 and momentum 0.9. The learning rate is decreased by a factor of 2 whenever the validation error does not improve for 20 epochs. The batch size is 256 and a maximum of 100k minibatch steps are used. 
We run all experiments, as explicitly stated, without data augmentation or with mild augmentation consisting of horizontal flipping and random cropping (in which the size is increased to $248\times 248$ before a crop of size  $224\times 224$ is selected).
Such fixed hyperparameters are suboptimal for some combinations of architectures and
datasets and therefore the resulting numbers can be worse than
state-of-the-art results. However, they enable reaching 100\% training accuracy in all combinations of models and datasets. For example, increasing the learning rate to 0.01 would prevent ResNet50 from progressing its training error beyond that of random guessing on the {\sc cassava} dataset.

\begin{table*}[tbp]
    \centering
    \small
    \vspace{-1.5ex}
    \caption{Significance analysis: A star ($\star$) in a cell indicates that the method in column outperforms the method in row with statistically significant evidence at the $95\%$ confidence level, computed using the exact binomial test. A circle ($\circ$) indicates that the significance test continues to hold even after applying Holm's step-down correction for multiple hypothesis tests \cite{demvsar2006statistical}. There is no architecture where an algorithm outperforms \reinit statistically significantly, so the last (empty) row is omitted.}
    \vspace{-1ex}
    \label{tab:stats_sig_results}
    \begin{tabularx}{2\columnwidth}{l|XXXXXX|XXXXXX|XXXXXX|XXXXXX}
    \toprule
      & \multicolumn{6}{c}{\texttt{scnn}} &  \multicolumn{6}{c}{\texttt{vgg16}} &  \multicolumn{6}{c}{\texttt{mobilenet}}&  \multicolumn{6}{c}{\texttt{resnet50}}\\\hline
      &{\sc b} & {\sc r} & {\sc d} & {\sc w} & {\sc f} & {\sc l}
      &{\sc b} & {\sc r} & {\sc d} & {\sc w} & {\sc f} & {\sc l}
      &{\sc b} & {\sc r} & {\sc d} & {\sc w} & {\sc f} & {\sc l}
      &{\sc b} & {\sc r} & {\sc d} & {\sc w} & {\sc f} & {\sc l}
      \\\hline
          {\sc b}  
            &&&&&$\circ$ &$\circ$ &
            &$\circ$ &&$\circ$ &$\circ$ &$\circ$ &
            &&$\circ$ &$\star$&&$\circ$ &
            &$\circ$ &$\circ$ &$\circ$ &$\circ$ &$\circ$ 
          \\
          {\sc r}
            &&&&&$\circ$ &$\circ$ &
            &&&&&$\circ$ &
            &&&&&$\circ$ &
            && &$\circ$ &&$\star$ 
          \\
          {\sc d} 
            &&$\star$ &&&$\circ$ &$\circ$ &
            &$\circ$ &&$\star$ &$\circ$ &$\circ$ &
            &&&&&$\circ$ &
            &&&&&
          \\
          {\sc w} 
            &&&&&$\circ$ &$\circ$ &
            &&&&$\star$ &$\circ$ &
            &&&&&$\circ$ &
            &&&&&
          \\
          {\sc f} 
          &&&&&&&
          &&&&&&
          &&&&&$\circ$ &
          &&&$\circ$ &&
          \\
           \bottomrule    \end{tabularx}
\end{table*}

Table~\ref{tab:reinit_results_with_aug} 
provides the detailed results of the five reinitialization methods across the benchmark datasets with augmentation. Appendix~\ref{appendix::detailed_figures} includes detailed results when other experiment settings are used.
We now re-run these experiments with varying settings including augmentation (with/without), dropout rate (0 or 0.25) \citep{srivastava2014dropout}, and initializer (He normal \citep{he2015delving} or Xavier uniform \citep{glorot2010understanding}) and then aggregate the observations regarding which reinitialization method performs better than others (and the baseline). We perform an exact binomial test to evaluate which method performs statistically significantly better across the settings. 
In Table~\ref{tab:stats_sig_results}, we summarize these results. We observe that all reinitialization methods perform better than the baseline. In addition, \reinit outperforms the other methods with a statistical significant evidence at the 95\% confidence level. Moreover, \fc performs generally better than \welsns, \welsrns, and \dsdns. It is worth reiterating, that \fc is a special case of \reinit that corresponds to $K=1$ and $N>1$.

\begin{figure}[tbp]
    \centering
    \includegraphics[height=0.4\columnwidth,width=0.4\columnwidth]{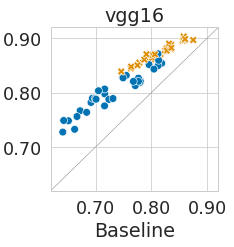}
    \includegraphics[height=0.4\columnwidth,width=0.4\columnwidth]{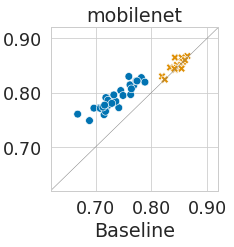} \hfill
    \includegraphics[height=0.4\columnwidth,width=0.4\columnwidth]{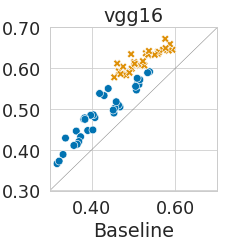}
    \includegraphics[height=0.4\columnwidth,width=0.4\columnwidth]{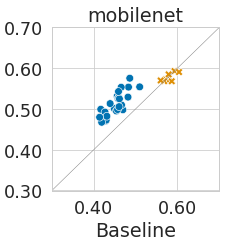}
    \caption{The test accuracy is displayed for the baseline ($x$-axis) vs.\ \reinit ($y$-axis) with $N=1$ for both \texttt{vgg16} and \texttt{mobilenet} on CIFAR10 (top) and CIFAR100 (bottom). Amber dots are for experiments with augmentation while blue dots correspond to experiments without augmentation.}
    \label{fig:to-be-added}
\end{figure}

In Figure~\ref{fig:to-be-added} we highlight a different view of the space of 
hyperparameters and reinitialization: For the architectures \texttt{vgg16} and \texttt{mobilenet} and the popular datasets CIFAR-10/100, we plot the results for a wider range of hyperparameters and compare the resulting accuracies for the baseline and \reinitns. Example of hyperparameters that we vary include the learning rate ($0.001$, $0.003$,  or $0.01$), dropout rate ($0$, $0.25$, or $0.5$), momentum ($0$ or $0.9$), weight decay ($0$ or $10^{-5}$), and initializer (He normal or Xavier uniform). We observe that \reinit consistently outperforms the baseline. Further, the resulting accuracies range up to results that are as expected in the `vanilla' setting we are considering here (without heavily tuning e.g.\ augmentation and regularization). Of course, better results can still be achieved when aiming for state-of-the-art results, see e.g.~\cite{sam2021} for an in-depth discussion of the current best results on the CIFAR-10/100.

\subsection{Effect of Experiment Design}
To determine when a particular reinitialization method outperforms others, we train a decision tree classifier on the outcomes of several experiments that vary in design by, for example, the choice of the initializer (He normal or Xavier uniform), augmentation, and dropout. Every setting contains experiment runs of each of the 5 reinitialization methods in addition to the baseline for the four architectures and 12 benchmark datasets. 

\begin{figure*}[tbp]\centering
\begin{tikzpicture}[node distance = 2cm, auto, scale=0.75, every node/.style={transform shape}]
    \node [block] (top) {ResNet?};
    \node [below of=top, node distance=1.5cm] (belowtop) {};
    \node [block, left of=belowtop, node distance=3cm] (l10) {MobileNet?};
    \node [block, right of=belowtop, node distance=3cm] (l11) {Training Set Size $<$ 35K?};

    \path [line,dashed] (top) -| node [near start] {No} (l10);
    \path [line,dashed] (top) -| node [near start] {Yes} (l11);

    \node [below of=belowtop, node distance=2cm] (belowtop2) {};
    \node [block, left of=belowtop2, node distance=6cm] (l20) {Augment?};
    \node [block, left of=belowtop2, node distance=3cm] (l21) {Training Set Size $>$ 35K?};

    \node [decision, below of=l11, node distance=3cm] (l22) {BL, DSD, WELS, FC, LW};
    \node [decision, right of=l22, node distance=3cm] (l23) {LW};

    \path [line,dashed] (l11) -- node {No}(l22);
    \path [line,dashed] (l11) -| node {Yes} (l23);

    \path [line,dashed] (l10) -| node [near start] {No}(l20);
    \path [line,dashed] (l10) -- node  {Yes} (l21);
    
    \node [decision, below of=l21, node distance=2.5cm] (l30) {LW};
    \node [decision, right of=l30, node distance=3cm] (l31) {LW, DSD};

    \path [line,dashed] (l21) -- node {No}(l30);
    \path [line,dashed] (l21) -| node {Yes} (l31);

    \node [decision, below of=l20, node distance=2.5cm] (140) {LW};

    \path [line,dashed] (l20) -- node  {Yes}(140);

    \node [block, left of=l20, node distance=3cm] (150) {Number of Classes $>$ 20?};
    \path [line,dashed] (l20) -- node  {No}(150);

    \node [decision, below of=150, node distance=2.5cm] (160) {LW, FC};

    \node [decision, left of=160, node distance=3 cm] (170) {LW, WELS};

    \path [line,dashed] (150) -- node  {Yes}(160);
    \path [line,dashed] (150) -| node [near start] {No} (170);
\end{tikzpicture}
\caption{A decision tree classifier trained to predict the best reinitialization method based on the experiment design. The features are the training set size, number of classes, neural network architecture, dropout rate, augmentation, and the choice of initializer. When node impurity is large, i.e.\ Gini index $> 0.5$, we report the top performing methods in the corresponding leaf.}\label{fig:dt}
\end{figure*}
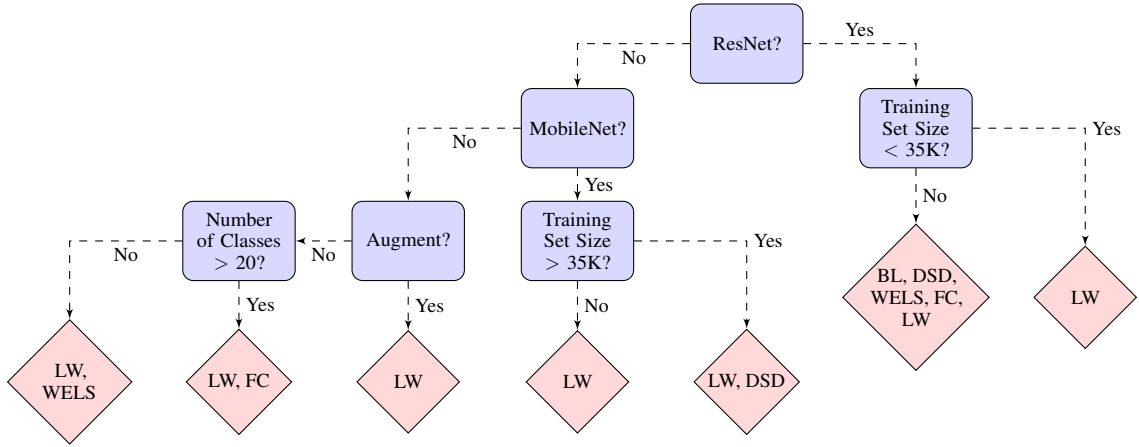
\begin{figure*}[tbp]
    \centering
    \includegraphics[width=1.95\columnwidth]{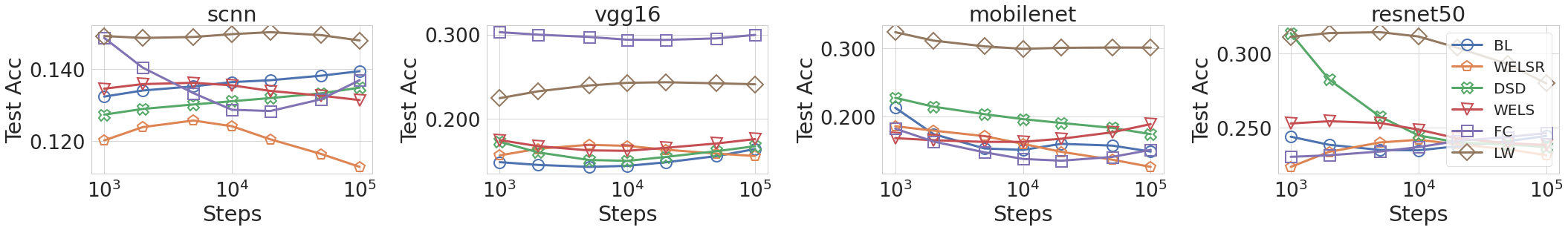}\\
    \caption{The test accuracy of reinitialization methods with different compute budgets plotted for Oxford-IIIT dataset (see Appendix \ref{appendix::full_figures} for full results on all datasets). The $x$-axis is the number of training steps per reinitialization round. For the baseline, the test accuracy is plotted over the same total number of steps as reinitialization. Most reinitialization methods quickly surpass the accuracy of the baseline for the same amount of compute.
    }
    \label{fig:reinit_compute}
\end{figure*}
\begin{figure*}[tbp]
    \centering
    \includegraphics[width=1.95\columnwidth]{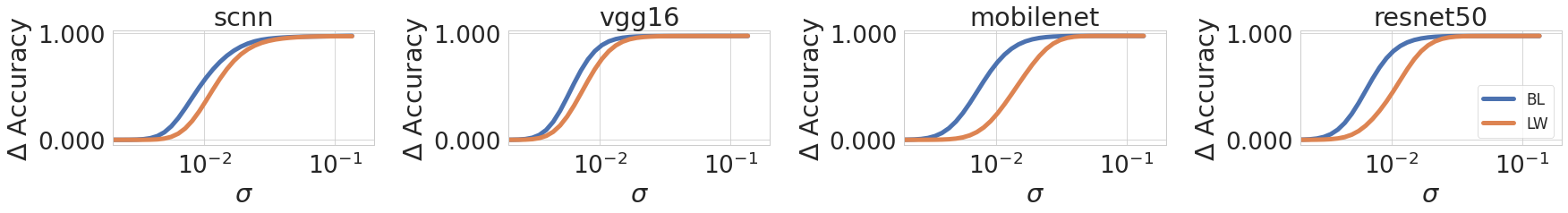}\\
    \caption{Bi-criteria plots for the change in training accuracy ($y$-axis) when the model parameters are perturbed by standard Gaussian noise $\mathcal{N}(0, \sigma^2I)$ for the Oxford-IIIT dataset. Lower curves suggest flatter local minima and better generalization. See Appendix \ref{appendix::full_figures} for full results on all datasets.}
    \label{fig:kl_acc}
\end{figure*}

\begin{figure*}[tbp]
    \centering
    \includegraphics[width=1.95\columnwidth]{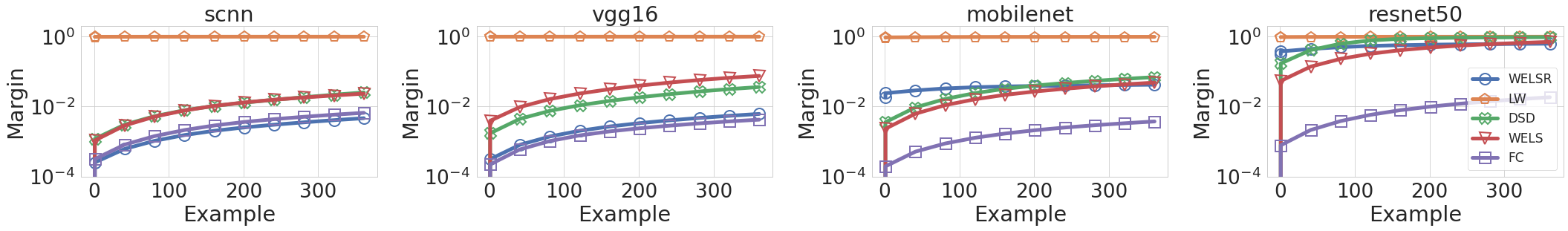}\\
    \includegraphics[width=1.95\columnwidth]{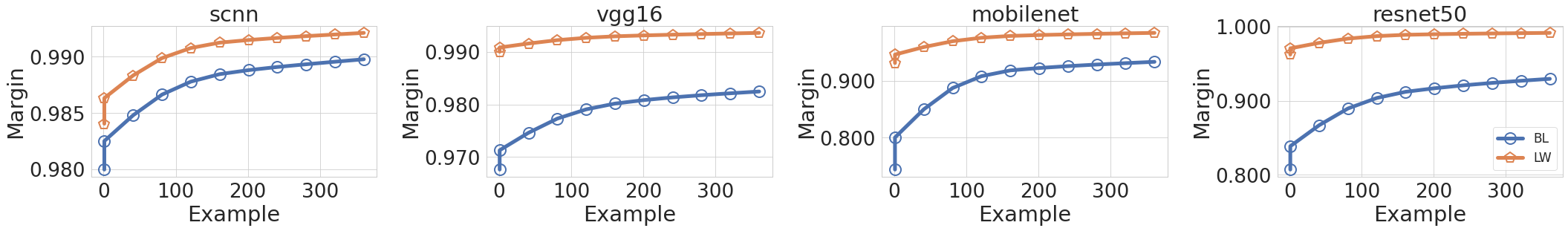}\\
    \caption{{\sc top:} The smallest 400 margins in the training sample are plotted for different reinitialization methods in the Oxford-IIIT dataset. \reinit (orange) boosts the margin considerably compared to previous reinitialization methods. {\sc bottom:} A comparison of the 400 smallest margins in the training sample between \reinit and {\sc bl}. The curves are displayed separately for a better visualization, as they almost coincide in the wide ranged log-scale in the top two rows. Appendix \ref{appendix::full_figures} contains full results on all datasets.}
    \label{fig:margins}
\end{figure*}
\subsection{Sharpness of the Local Minima}

We use the decision tree classifier, implemented using the Scikit-Learn package~\cite{scikit-learn}, for interpretability. The goal is to predict which reinitialization method performs best. Two features related to the dataset are included: the training set size and the number of classes. We use a minimum leaf size of 7 in the decision tree and a maximum depth of 4. Figure~\ref{fig:dt} displays the resulting decision tree. First, we observe that \reinit  improves performance across the majority of  combinations. The only exception is using ResNet50 with large training sets, in which most methods perform as well as the baseline.

\subsection{Compute}
In Table~\ref{tab:reinit_results_with_aug}, every reinitialization round is trained until convergence. However, improvements in generalization can also be obtained at lower computational overhead by stopping earlier than at convergence in each round. This is illustrated in Figure~\ref{fig:reinit_compute}. As shown in the figure, stopping early per round allows to realize the gain of reinitialization without incurring significant additional overhead. In addition, we show in Appendix \ref{appendix::training_speed} that training is faster in subsequent rounds of reinitialization.

\section{Analysis}\label{sect::analysis}
\subsection{Boosting the Margin}
As discussed earlier in Section~\ref{sect::related_work}, a typical approach for bounding the generalization gap in deep neural networks is to use a particular measure of the {size of the weights} normalized by the \emph{margin} on the training sample. Let $D$ be the number of layers in a neural network, whose output is a composition of functions: $f(x) = f_1\circ f_2\circ \cdots f_D(x)$, 
where each $f_i(x)$ is of the form $f_i(x)=\sigma(W_ix)$ for some matrix $W_i$ and ReLU activation $\sigma$. Then, one measure of the size of the weight that relates to generalization is the product of the Frobenius norms of layers $\prod_{i=1}^d||W||_F^2$~\citep{neyshabur2015norm,neyshabur2017exploring}. This is normalized by the margin $\gamma>0$ on the training examples, which is the smallest difference between the score assigned to the true label and the next largest score. For a better visualization, we use the margin of the softmax output since it is normalized in the interval $[0,\,1]$.

Figure~\ref{fig:margins} displays the smallest 400 margins on the training sample for Oxford-IIIT dataset. Appendix \ref{appendix::full_figures} contains full results on all datasets. As shown in the figure, \reinit boosts the margin on the training sample considerably when compared to previous reinitialization methods. Most importantly, \reinit achieves this \emph{without} increasing the size of the weights. To take the contribution of the normalization layers into account when calculating the product $\prod_{i=1}^d||W||_F^2$, we compare the product of the norms of the input to the classifier head (activations) and the norm of the weights of the classifier head in each method. We observe that \reinit tends to maintain the same size of the weights as the baseline. Appendix~\ref{appendix::size_w} provides further details. 

We provide an informal argument for why this happens. First, the product of the norms of the weights in the identified $K$ blocks in \reinit (cf.\ Figure~\ref{fig:reinit} and Algorithm~\ref{algorithm}) tend to remain unchanged due to the normalization layers inserted after each round. What changes is the norm of the \emph{final} layers (following block $K$), but their norm tends to shrink because they train from scratch faster with each round (cf. Appendix~\ref{appendix::training_speed}). As for the margin, because the network classifies all examples correctly in a few epochs in the final round of \reinitns, any additional epochs have the effect of increasing the margin to reduce the cross entropy loss. 

\subsection{Sharpness of the Local Minima}
Finally, we observe that the final solution provided by \reinit seems to reside in a ``flatter'' local minima of the loss surface than in the baseline. One method for quantifying flatness is to compare the impact on the training loss when the model parameters are perturbed by some standard Gaussian noise, which can be linked to generalization via explicit bounds \citep{neyshabur2017exploring}. To recall, both \reinit and {\sc bl} share the same size of the weights (cf. Appendix \ref{appendix::size_w}).  Figure~\ref{fig:kl_acc} shows that the solution reached by \reinit is more robust to model perturbation than in standard training. More precisely, for every amount of noise added into the model parameters \textbf{w}, the change in the training loss in \reinit is smaller than in standard training suggesting that the local minimum is flatter in \reinitns.

\section{Discussion}\label{sect::discussion}
In this paper, we present a comprehensive evaluation of reinitialization algorithms and introduce a new method that outperforms others.  Empirical results show that this method improves generalization across a wide range of architectures and hyper-parameters, particularly for small datasets. It relates  to prior works that distinguish learning general rules in earlier layers from exceptions to the rules in later layers, because \reinit places more emphasis on the early layers of the neural network. We also argue that the improved generalization can be connected to the sharpness of the local minima and the margins on the training examples, and conducted further ablation and failure analyses using decision trees. 

Our takeaway message is that the accuracy of convolutional neural networks can be improved for small datasets using bottom-up layerwise reinitialization, where the number of reinitialized layers may vary depending on the available compute budget. At one extreme, one would benefit from reinitializing the classifier's head alone, but reinitializing all layers in sequence with rescaling and normalization yields better results. For large datasets, however, reinitialization does not seem to offer a benefit.
We hope that the description of the observed positive effects will inspire others to study them  more and to develop more efficient alternatives.

\clearpage
\section*{Acknowledgement}
The authors would like to thank Ilya Tolstikhin, Robert Baldock, Behnam Neyshabur, Hanie Sedghi, Chiyuan Zhang, Hugo Larochelle and Mike Mozer for the useful discussions as well as the Google Brain Team at large for their support.

\bibliography{reinit}

\clearpage

\appendix

\section{Synthetic Data Experiment}\label{sect::appendix_toy_example}
We use the same type of data as described in Section~\ref{sect::syntheticdata}, but look in more 
detail at the more difficult case $\alpha=0.5$, this means the first three entries of the data encode 
the 8 possible labels as the 8 corners of the cube $[-0.5,0.5]^3$, whereas the remaining 
entries are still sampled from the standard normal distribution. In addition, one may add a weight decay penalty to the task and examine the impact of rescaling alone. Specifically, we consider two cases:
\begin{itemize}
    \item \emph{Rescaling}:
    Instead of training once for $N$ epochs, we train 5 times for $N/5$ epochs, and in between we scale back
    all weights such that the norm of each layer matches the norm after initialization.
    \item \emph{Reinitialization}:
    In addition to rescaling, we re-initialize the layers above the first one in the first two rounds,
    above the second layer in the next two rounds, and only the top layer in the last round.
\end{itemize}

The results are shown in Table~\ref{tab:TestAccRegularized}.  We use the same type of data as described above, but focus now at the more difficult case of $\alpha=0.5$. 

We observe that one can get significantly better results with weight decay. Nevertheless, \reinit gives an additional 
benefit on top of the L2 regularization. In this particular experiment, rescaling seems to have the biggest effect but this is not generally the case in natural image datasets, in which the gain seems to be modest without reinitialization.

\begin{table}[h]
    \centering
    \small
    \caption{Test accuracies (average of 100 runs)}
    \label{tab:TestAccRegularized}
    \begin{tabularx}{0.99\columnwidth}{XXXX}
    \toprule
      L2 penalty & Baseline & Rescaling & \reinit\\
      \midrule
         0.0\phantom{00}  &  0.19  &  0.21 & 0.25 \\
         0.005 & 0.51  &  0.67 & 0.82\\
         0.01\phantom{0}  & 0.54  &  0.89 & 0.85\\
         0.02\phantom{0}  & 0.58  &  0.87 & 0.86\\
         0.05\phantom{0}  & 0.77  &  0.78 & 0.79 \\
         0.1\phantom{00}   & 0.59  &  0.63 & 0.64 \\
      \bottomrule
    \end{tabularx}
\end{table}

\section{Experiment Setup}\label{appendix::exp_setup}
\subsection{Architectures}
Throughout the main text, we use four different architectures: one simple convolutional neural network, and three standard deep convolutional models. 

In all architectures, we use weight decay with penalty $10^{-5}$ unless explicitly stated otherwise. We also use layer normalization \citep{ba2016layer}, implemented in TensorFlow \citep{abadi2016tensorflow} using \texttt{GroupNormalization} layers with \texttt{groups=1}. Similar results are obtained when using Batch Normalization \citep{ioffe2015batch}.

In all experiments, we use SGD as an optimizer. Unless explicitly stated otherwise, we use a learning rate of 0.003 and momentum 0.9. Also, we use a batch size of 256. All experiments are executed on Tensor Processing Units 
for a maximum of 100,000 minibatch steps per reinitialization round. We resize images to  $224\times 224$ in all experiments.

\paragraph{Simple CNN (\texttt{scnn}).}This architecture contains four convolutional blocks followed by one dense layer before the classifier head. The number of convolutional blocks $K$ used in this architecture is 4. 
Every convolutional block is a 2D convolutional layer, followed by layer normalization and ReLU activation. Precisely:
\begin{footnotesize}
\begin{verbatim}
    conv2d              32 filters
    layer_norm; 
    activation_relu
    
    conv2d              32 filters
    layer_norm; 
    activation_relu
    max_pooling2d
    
    conv2d              64 filters
    layer_norm; 
    activation_relu
    
    conv2d              64 filters
    layer_norm; 
    activation_relu
    max_pooling2d
    
    flatten
    dense              512 units
    layer_norm; 
    activation_relu
    dropout
    
    classifier_head
\end{verbatim}
\end{footnotesize}

\paragraph{MobileNetV1 (\texttt{mobilenet}).}
This is the standard shallow MobileNet architecture \citep{howard2017mobilenets}. The standard blocks in this architecture are either convolutional blocks with layer normalization and ReLU or depthwise separable convolutions with depthwise and pointwise layers followed by layer normalization and ReLU (see Figure 3 in \citet{howard2017mobilenets}). In the shallow architecture, the number of convolutional blocks $K$ is 7.

\paragraph{VGG16 (\texttt{vgg16}).}
This is the standard VGG16 architecture \citep{simonyan2014very}. The standard blocks in this architecture are convolutional layers with layer normalization and ReLU (see Table 1 in \citet{simonyan2014very}). The number of convolutional blocks $K$ is 13.

\paragraph{ResNet50 (\texttt{resnet50}).}
This is the standard ResNet50 architecture \citep{he2016deep}. The standard blocks in this architecture either identity blocks or convolutional blocks (see Table 1 in \citet{he2016deep}). The number of convolutional blocks $K$ used in this architecture is 16.

\subsection{Datasets}
The 12 benchmark datasets are all taken from the Tensorflow dataset repository \citep{abadi2016tensorflow}. Table \ref{tab:datasets} gives an overview of each dataset.

\begin{table}[h]
    \centering
    \small
    \caption{Overview of the 12 benchmark datasets.}
    \label{tab:datasets}
    \vspace{1ex}
    \begin{tabularx}{\columnwidth}{@{}XXXX@{}}
    \toprule
      Name & |Training| & |Test| & \# Classes\\
      \midrule
        {\sc oxford-iiit}&\phantom{0}3,680 &\phantom{0}3,669 &\phantom{0}37 \\
        \multicolumn{4}{@{}l}{\cite{parkhi12a}}\\
        \midrule
        {\sc dogs} &12,000 &\phantom{0}8,580 &120 \\
        \multicolumn{4}{@{}l}{\cite{KhoslaYaoJayadevaprakashFeiFei_FGVC2011}}\\
        \midrule
        {\sc fmnist}  &60,000 &10,000 &\phantom{0}10 \\
        \multicolumn{4}{@{}l}{\cite{DBLP:journals/corr/abs-1708-07747}}\\
        \midrule
        {\sc cars196} &\phantom{0}8,144 &\phantom{0}8,041 &196  \\
        \multicolumn{4}{@{}l}{\cite{KrauseStarkDengFei-Fei_3DRR2013}}\\
        \midrule
        
        {\sc mnist-cor} &60,000 &10,000 &\phantom{0}10 \\
        \multicolumn{4}{@{}l}{\cite{mu2019mnist}}\\
        \midrule
        {\sc cifar10-cor} &50,000 &10,000 &\phantom{0}10 \\
        \multicolumn{4}{@{}l}{\cite{hendrycks2018benchmarking}}\\
        \midrule
        {\sc cifar10} &50,000 &10,000 &\phantom{0}10 \\
        \multicolumn{4}{@{}l}{\cite{Krizhevsky09learningmultiple}}\\
        \midrule
        {\sc caltech101} &\phantom{0}3,060 &\phantom{0}6,084 & 101\\
        \multicolumn{4}{@{}l}{\cite{FeiFei2004LearningGV}}\\
        \midrule
        
        {\sc cassava} &\phantom{0}5,656 &\phantom{0}1,885 & \phantom{00}4 \\
        \multicolumn{4}{@{}l}{\cite{mwebaze2019icassava}}\\
        \midrule
        {\sc cmaterdb} &\phantom{0}5,000 &\phantom{0}1,000 &\phantom{0}10\\
        \multicolumn{4}{@{}l}{\cite{Das:2012:SFC:2240301.2240421}}\\
        \midrule
        {\sc birds2010}&\phantom{0}3,000 &\phantom{0}3,033 &200 \\
        \multicolumn{4}{@{}l}{\cite{WelinderEtal2010}}\\
        \midrule
        {\sc cifar100}&50,000 &10,000 &100 \\
        \multicolumn{4}{@{}l}{\cite{Krizhevsky09learningmultiple}}\\
      \bottomrule
    \end{tabularx}
\end{table}

\section{Detailed Empirical Evaluation Results} \label{appendix::detailed_figures}
As stated earlier in Section \ref{sect::experiments}, Table \ref{tab:reinit_results_with_aug} provides the results when augmentation is used but without dropout. We provide here two other settings: without augmentation and dropout and with both augmentation and dropout. We use the same other hyperparameters described in Section \ref{sect::experiments}. Table \ref{tab:reinit_results_no_aug} provides the results without augmentation and without dropout. Table \ref{tab:reinit_results_aug_dropout} provides the results with both augmentation and dropout. We did not run the combination of dropout without augmentation.

\begin{table}[tbp]
    \centering
    \footnotesize
    \caption{Test accuracy results [\%] for the five reinitialization methods across 12 benchmark datasets. All reinitialization methods improve generalization, with \reinit outperforming others in most cases. Here no data augmentation or dropout are used, see Table \ref{tab:reinit_results_with_aug} for results with data augmentation and Table \ref{tab:reinit_results_aug_dropout} for results with both augmentation and dropout.}
    \label{tab:reinit_results_no_aug}
    \vspace{1ex}
    \begin{tabularx}{\columnwidth}{@{}llXXXXXX@{\hspace{0.95em}}}
    \toprule
      \em Dataset& \em Model & \sc b & \sc r &  \sc d & \sc w & \sc f &\sc l\\\midrule
\multirow{4}{*}{\sc oxford-iiit}&\texttt{scnn} &13.7 &12.0 &13.4 &14.0 &14.0 &\bf15.0 \\&\texttt{vgg16} &16.2 &16.7 &16.9 &16.4 &\bf29.7 &25.6 \\&\texttt{mobilenet} &13.4 &13.8 &17.2 &15.9 &14.8 &\bf30.1 \\&\texttt{resnet50} &24.0 &24.2 &22.9 &24.9 &23.7 &\bf28.5 \\\midrule
\multirow{4}{*}{\sc dogs}&\texttt{scnn} &\phantom{0}5.8 &\phantom{0}5.2 &\phantom{0}5.2 &\phantom{0}5.1 &\phantom{0}5.1 &\bf\phantom{0}6.1 \\&\texttt{vgg16} &11.4 &11.8 &10.7 &10.3 &\bf24.2 &19.1 \\&\texttt{mobilenet} &\phantom{0}9.7 &\phantom{0}8.6 &\phantom{0}9.0 &11.6 &\phantom{0}9.9 &\bf19.2 \\&\texttt{resnet50} &14.0 &17.5 &19.2 &16.0 &13.9 &\bf22.0 \\\midrule
\multirow{4}{*}{\sc fmnist}&\texttt{scnn} &91.8 &91.7 &91.7 &91.7 &91.8 &\bf92.1 \\&\texttt{vgg16} &92.4 &92.6 &92.4 &92.6 &\bf92.7 &92.5 \\&\texttt{mobilenet} &92.2 &91.7 &91.8 &91.9 &91.9 &\bf92.6 \\&\texttt{resnet50} &89.9 &89.7 &89.9 &90.0 &90.1 &\bf90.2 \\\midrule
\multirow{4}{*}{\sc cars196}&\texttt{scnn} &\phantom{0}3.4 &\phantom{0}3.4 &\phantom{0}3.4 &\phantom{0}3.5 &\phantom{0}3.2 &\bf\phantom{0}3.8 \\&\texttt{vgg16} &\phantom{0}6.5 &\phantom{0}5.5 &\phantom{0}7.7 &\phantom{0}7.0 &\bf20.9 &\phantom{0}9.9 \\&\texttt{mobilenet} &\phantom{0}6.1 &\phantom{0}8.9 &\phantom{0}5.3 &\phantom{0}7.4 &\phantom{0}6.8 &\bf22.2 \\&\texttt{resnet50} &10.1 &12.8 &13.4 &\bf13.7 &10.4 &12.6 \\\midrule
\multirow{4}{*}{\sc mnist-cor}&\texttt{scnn} &99.2 &99.3 &99.3 &99.3 &\bf99.4 &\bf99.4 \\&\texttt{vgg16} &99.3 &99.3 &99.4 &99.4 &99.4 &\bf99.5 \\&\texttt{mobilenet} &\bf99.5 &99.4 &99.4 &99.3 &99.4 &\bf99.5 \\&\texttt{resnet50} &99.0 &99.1 &99.0 &99.1 &\bf99.2 &99.0 \\\midrule
\multirow{4}{*}{\sc cifar10-cor}&\texttt{scnn} &71.8 &75.1 &73.7 &\bf75.7 &74.9 &75.2 \\&\texttt{vgg16} &76.7 &79.2 &78.5 &80.5 &77.7 &\bf81.6 \\&\texttt{mobilenet} &78.3 &78.4 &78.0 &78.4 &79.2 &\bf83.6 \\&\texttt{resnet50} &68.4 &69.2 &\bf71.3 &70.8 &69.4 &69.7 \\\midrule
\multirow{4}{*}{\sc cifar10}&\texttt{scnn} &72.2 &75.1 &74.5 &\bf75.9 &74.8 &\bf75.9 \\&\texttt{vgg16} &76.8 &79.3 &78.6 &80.8 &77.4 &\bf81.4 \\&\texttt{mobilenet} &77.7 &77.7 &78.7 &79.8 &79.3 &\bf83.6 \\&\texttt{resnet50} &68.8 &70.4 &\bf71.2 &71.0 &69.2 &70.2 \\\midrule
\multirow{4}{*}{\sc caltech101}&\texttt{scnn} &\bf51.1 &49.9 &49.8 &48.4 &50.8 &50.7 \\&\texttt{vgg16} &54.1 &55.4 &54.2 &55.9 &\bf69.1 &57.3 \\&\texttt{mobilenet} &36.7 &43.0 &44.9 &40.9 &42.3 &\bf47.0 \\&\texttt{resnet50} &50.4 &54.6 &\bf55.0 &52.9 &50.7 &53.3 \\\midrule
\multirow{4}{*}{\sc cassava}&\texttt{scnn} &59.4 &58.6 &58.6 &59.8 &61.6 &\bf63.9 \\&\texttt{vgg16} &58.4 &59.5 &58.4 &\bf62.1 &58.5 &59.3 \\&\texttt{mobilenet} &52.6 &57.9 &63.6 &62.2 &55.4 &\bf70.0 \\&\texttt{resnet50} &61.9 &57.3 &62.8 &58.1 &56.7 &\bf63.9 \\\midrule
\multirow{4}{*}{\sc cmaterdb}&\texttt{scnn} &97.1 &97.0 &96.9 &97.3 &\bf97.4 &\bf97.4 \\&\texttt{vgg16} &97.8 &97.2 &97.5 &97.5 &97.7 &\bf99.0 \\&\texttt{mobilenet} &\bf98.6 &\bf98.6 &\bf98.6 &98.6 &98.1 &\bf98.6 \\&\texttt{resnet50} &94.5 &94.9 &\bf97.1 &95.9 &96.7 &95.8 \\\midrule
\multirow{4}{*}{\sc birds2010}&\texttt{scnn} &\phantom{0}2.0 &\bf\phantom{0}2.6 &\phantom{0}2.3 &\phantom{0}2.2 &\phantom{0}2.4 &\phantom{0}2.4 \\&\texttt{vgg16} &\phantom{0}3.8 &\phantom{0}4.3 &\phantom{0}4.1 &\phantom{0}3.4 &\bf\phantom{0}8.5 &\phantom{0}5.1 \\&\texttt{mobilenet} &\phantom{0}4.5 &\phantom{0}3.9 &\phantom{0}5.2 &\phantom{0}5.9 &\phantom{0}5.3 &\bf\phantom{0}8.1 \\&\texttt{resnet50} &\phantom{0}6.9 &\phantom{0}8.7 &\bf10.0 &\bf10.0 &\phantom{0}6.5 &\bf10.0 \\\midrule
\multirow{4}{*}{\sc cifar100}&\texttt{scnn} &39.8 &42.2 &39.5 &40.5 &43.7 &\bf43.8 \\&\texttt{vgg16} &45.2 &48.4 &47.9 &\bf51.6 &45.2 &51.2 \\&\texttt{mobilenet} &44.7 &46.1 &\bf54.3 &44.3 &42.0 &50.9 \\&\texttt{resnet50} &36.4 &37.5 &\bf40.4 &38.9 &36.7 &38.0\\ 
\bottomrule
    \end{tabularx}
\end{table}

\begin{table}[tbp]
    \centering
    \footnotesize
    \caption{Test accuracy results [\%] for the five reinitialization methods across 12 benchmark datasets. All reinitialization methods improve generalization, with \reinit outperforming others in most cases. Here both data augmentation and dropout are used, see Table \ref{tab:reinit_results_with_aug} for results with data augmentation but no dropout and Table \ref{tab:reinit_results_no_aug} for results without augmentation or dropout.}
    \vspace{1ex}
    \label{tab:reinit_results_aug_dropout}
    \begin{tabularx}{\columnwidth}{@{}llXXXXXX@{\hspace{0.95em}}}
    \toprule
      \em Dataset& \em Model & \sc b & \sc r &  \sc d & \sc w & \sc f &\sc l\\\midrule

\multirow{4}{*}{\sc oxford-iiit}&\texttt{scnn} &15.8  &13.6  &14.3  &15.1  &\bf 19.3  &17.1  \\&\texttt{vgg16} &25.3  &26.7  &26.6  &26.4  &\bf 43.4  &34.4  \\&\texttt{mobilenet} &28.4  &29.1  &27.8  &27.9  &22.0  &\bf 41.0  \\&\texttt{resnet50} &33.1  &35.2  &\bf 41.2  &37.4  &32.6  &35.6  \\

\midrule\multirow{4}{*}{\sc dogs}&\texttt{scnn} &\phantom{0}8.4  &\phantom{0}8.9  &\phantom{0}7.6  &\phantom{0}8.0  &\bf \phantom{0}9.6  &\phantom{0}9.1  \\&\texttt{vgg16} &17.7  &19.7  &19.5  &18.9  &34.9  &\bf 35.9  \\&\texttt{mobilenet} &17.8  &23.5  &27.3  &22.4  &20.5  &\bf 35.0  \\&\texttt{resnet50} &30.7  &33.3  &33.1  &33.8  &34.5  &\bf 40.1  \\

\midrule\multirow{4}{*}{\sc fmnist}&\texttt{scnn} &93.1  &93.0  &92.8  &93.2  &\bf 93.3  &93.0  \\&\texttt{vgg16} &91.7  &92.4  &91.7  &92.1  &92.4  &\bf 93.0  \\&\texttt{mobilenet} &91.6  &91.7  &91.7  &92.1  &91.7  &\bf 92.9  \\&\texttt{resnet50} &93.1  &93.3  &93.2  &\bf 93.4  &92.9  &\bf 93.4  \\

\midrule\multirow{4}{*}{\sc cars196}&\texttt{scnn} &\phantom{0}6.3  &\phantom{0}6.2  &\phantom{0}5.9  &\phantom{0}6.8  &\bf\phantom{0}7.4  &\phantom{0}6.4  \\&\texttt{vgg16} &14.3  &18.9  &12.6  &16.6  &\bf 45.2  &34.2  \\&\texttt{mobilenet} &\phantom{0}9.5  &16.1  &30.8  &24.1  &16.4  &\bf 44.5  \\&\texttt{resnet50} &19.8  &45.7  &43.0  &\bf 48.1  &45.0  &47.5  \\\midrule

\multirow{4}{*}{\sc mnist-cor}&\texttt{scnn} &99.0  &99.0  &99.1  &98.9  &\bf 99.2  &99.0  \\&\texttt{vgg16} &98.9  &99.0  &99.0  &99.0  &99.0  &\bf 99.2  \\&\texttt{mobilenet} &98.8  &\bf 99.0  &\bf 99.0  &98.9  &\bf 99.0  &\bf 99.0  \\&\texttt{resnet50} &\bf 99.1  &\bf 99.1  &\bf 99.1  &\bf 99.1  &\bf 99.1  &\bf 99.1  \\

\midrule\multirow{4}{*}{\sc cifar10-cor}&\texttt{scnn} &84.0  &85.3  &84.9  &85.0  &85.2  &\bf 85.8  \\&\texttt{vgg16} &84.8  &86.5  &84.9  &85.9  &85.4  &\bf 88.8  \\&\texttt{mobilenet} &85.9  &87.2  &88.2  &\bf 88.3  &85.8  &87.0  \\&\texttt{resnet50} &89.5  &89.6  &\bf 90.7  &90.0  &90.3  &89.1  \\

\midrule\multirow{4}{*}{\sc cifar10}&\texttt{scnn} &84.6  &\bf 86.0  &85.8  &85.4  &85.9  &85.6  \\&\texttt{vgg16} &85.0  &87.1  &85.3  &85.6  &85.9  &\bf 89.5  \\&\texttt{mobilenet} &86.6  &86.9  &88.3  &\bf 88.5  &86.7  &87.4  \\&\texttt{resnet50} &89.6  &90.0  &90.7  &\bf 91.2  &89.7  &89.5  \\

\midrule\multirow{4}{*}{\sc caltech101}&\texttt{scnn} &51.4  &53.5  &51.1  &52.6  &52.4  &\bf 53.6  \\&\texttt{vgg16} &59.6  &61.7  &60.8  &61.0  &\bf 68.1  &62.7  \\&\texttt{mobilenet} &47.1  &43.0  &47.5  &45.6  &\bf 51.3  &49.4  \\&\texttt{resnet50} &50.6  &54.1  &54.6  &\bf 55.7  &51.4  &50.3  \\

\midrule\multirow{4}{*}{\sc cassava}&\texttt{scnn} &61.8  &62.9  &60.7  &61.9  &68.4  &\bf 68.8  \\&\texttt{vgg16} &70.1  &71.4  &\bf 73.7  &71.0  &71.1  &71.9  \\&\texttt{mobilenet} &68.5  &70.0  &78.6  &74.3  &73.7  &\bf 80.5  \\&\texttt{resnet50} &74.2  &77.6  &\bf 81.5  &80.9  &73.5  &78.6  \\

\midrule\multirow{4}{*}{\sc cmaterdb}&\texttt{scnn} &97.0  &97.4  &\bf 98.7  &97.7  &97.9  &97.7  \\&\texttt{vgg16} &97.9  &97.7  &97.9  &97.8  &\bf 98.3  &97.8  \\&\texttt{mobilenet} &97.4  &\bf 98.6  &97.9  &97.5  &97.3  &98.2  \\&\texttt{resnet50} &97.8  &97.5  &98.0  &97.8  &\bf 98.6  &\bf 98.6  \\

\midrule\multirow{4}{*}{\sc birds2010}&\texttt{scnn} &\phantom{0}4.0  &\phantom{0}4.0  &\phantom{0}3.5  &\phantom{0}3.7  &\phantom{0}3.3  &\bf \phantom{0}4.4  \\&\texttt{vgg16} &\phantom{0}6.5  &\phantom{0}8.1  &\phantom{0}8.5  &\phantom{0}8.1  &\bf 18.6  &12.6  \\&\texttt{mobilenet} &\phantom{0}2.5  &\bf \phantom{0}9.6  &\phantom{0}5.1  &\phantom{0}6.3  &\phantom{0}8.4  &\phantom{0}9.0  \\&\texttt{resnet50} &11.5  &14.8  &13.3  &\bf 15.2  &13.5  &13.3  \\

\midrule\multirow{4}{*}{\sc cifar100}&\texttt{scnn} &57.0  &56.9  &56.9  &58.3  &\bf 58.8  &\bf 58.8  \\&\texttt{vgg16} &56.7  &61.1  &57.2  &58.5  &57.4  &\bf 64.5  \\&\texttt{mobilenet} &58.0  &60.6  &60.4  &\bf 64.8  &58.2  &58.1  \\&\texttt{resnet50} &59.6  &61.4  &61.8  &\bf 63.5  &63.1  &62.2  \\

\bottomrule
    \end{tabularx}
\end{table}

\section{Size of the
Weights}\label{appendix::size_w}
\begin{figure*}[h]
    \centering
    \includegraphics[width=1.9\columnwidth]{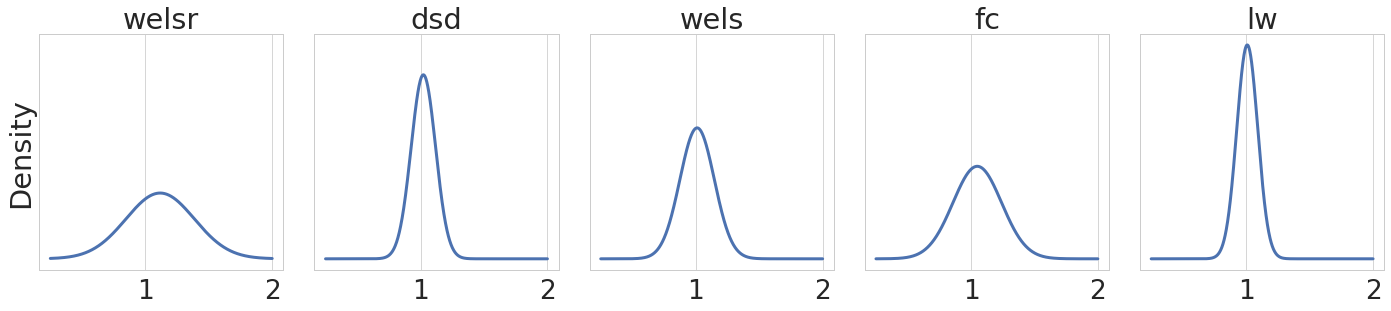}
    \caption{For each reinitialization method, the Gaussian approximation of the density of the \emph{ratio} of the size of the weights over the size of the weights in the baseline method is shown. The density of the ratio in \reinit is concentrated around 1, which implies that \reinit tends to not increase the size of the weights.}
    \label{fig:hist_w}
\end{figure*}

To calculate the norm of the weights while taking the contribution of the normalization layers into account, we compute the norm of the input to the classifier head (activations) for a random training sample of size 256. Then, we compute the Frobenius norm of the weights at the classifier head. Finally, we compute their product, which reflects the product of the Frobenius norm of layers stated in the generalization bound. Figure~\ref{fig:hist_w} shows a a Gaussian approximation to the ratio of the size of the weights of each reinitialization method over the size of the weights in the baseline. As shown in the figure, \reinit tends to maintain the size of the weights, while also boosting the margin on the training examples as discussed in Section~\ref{sect::analysis}

\section{Training Speed}\label{appendix::training_speed}

\begin{figure}[h]
    \centering
    \includegraphics[width=\columnwidth]{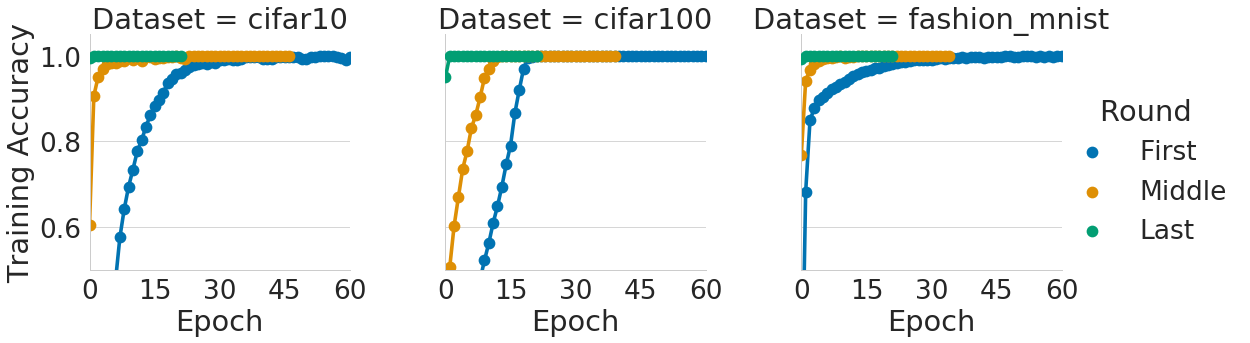} \vspace{-2ex}
    \caption{Training accuracy plotted against epochs in the first, middle, and last rounds of \reinit with  $N=1$ in  \texttt{vgg16} (see Appendix~\ref{appendix::training_speed}). A training round corresponds to fine-tuning the model until convergence. While \reinit involves multiple rounds of training, training can be much faster in subsequent rounds.}
    \label{fig:simplifiedCNN}
\end{figure}

It is worth highlighting that while \reinit involves multiple rounds of training the whole model, training is often much faster in subsequent rounds. This is illustrated in Figure~\ref{fig:simplifiedCNN} for the \texttt{vgg16} architecture. The experiment settings used to create Figure~\ref{fig:simplifiedCNN} are:

\begin{tabularx}{\columnwidth}{X@{\hspace{2em}}XX}
     \em Learning Rate &0.003  &\\
     \em Momentum &0 &\\
     \em Weight Decay &1e-5 &\\
     \em Augmentation &No &\\
     \em Dropout &0 &\\
     \em Initializer &He Normal &
\end{tabularx}

\section{Ablation}\label{appendix:ablation}
\reinit includes rescaling, normalization, and reinitialization. In some cases, these may not all be required and reinitialization alone suffices, but this is not always the case. We observe a consistent improvement in \reinit  when rescaling and normalization are included, in addition to fine-tuning the whole model at each round. In general:
\begin{itemize}
\item The improvement in generalization in \reinit cannot be attributed to rescaling or normalization alone. Reinitialization has the main effect.
\item There exist experiment designs in which reinitialization fails without normalization or fine-tuning the model. 
\item We observe cases in which rescaling alone helps but adding reinitialization improves performance further.
\item The gain from \reinit cannot be obtained by just training the baseline longer (i.e.\ using the same computational budget).
\end{itemize}

In this section, we show that the primary effect in \reinit comes from reinitialization, and that the improvement in generalization cannot be attributed to rescaling or normalization alone. We also show that fine-tuning the whole model performs better than freezing the early layers. Finally, we illustrate a case where \reinit without normalization fails.

\paragraph{Rescaling.}In the synthetic data experiment in Appendix~\ref{sect::appendix_toy_example}, we show that rescaling improves performance compared to the baseline but adding re-initializations improves it further. 

\paragraph{Reinitialization.} We use the \texttt{vgg16} architecture with the same hyperparameters as listed in Appendix~\ref{appendix::training_speed}.

We train it on CIFAR10 and CIFAR100. First, we observe that applying the same sequential process with rescaling and normalization but without reinitialization does not have an impact on the test accuracy. The test accuracy in CIFAR10 remains at around 66\% and in CIFAR100 at around 34\% in all rounds, similar to the baseline (this is different from the results in Table~\ref{tab:reinit_results_no_aug} because momentum is not used here). When reinitialization is added, we obtain the familiar looking curves where the test accuracy improves steadily with each round. In particular, it reaches around 75\% in CIFAR10 and around 42\% in CIFAR100. This shows that the improvement in \reinit cannot be attributed to normalization or rescaling alone.

\paragraph{Fine-tuning vs. Freezing.}
\reinit fine-tunes the entire model in each round. One alternative approach is to \emph{freeze} the early blocks. However, because of the co-adaptability  between neurons that arises during training~\cite{yosinski2014transferable}, freezing some layers and fine-tuning the rest is difficult to optimize and can harm its performance~\cite{yosinski2014transferable}. This is also true for reinitialization methods in general. Hence, the entire model including the kept layers is fine-tuned at each round.  

We illustrate this with one example. The architecture is \texttt{vgg16} on CIFAR100 but without normalization layers. If we apply \reinit while freezing the early layers instead of fine-tuning them, the training accuracy stays at 100\% after each round until about round 5 before it drops to 1\% and stays at 1\% training accuracy throughout the subsequent rounds. Fine-tuning the whole model does not exhibit this behavior. 

\paragraph{Normalization.}
\reinit inserts normalization layers after each round with no trainable parameters. To illustrate why normalization is important, let the architecture be \texttt{vgg16} on CIFAR100 again but without normalization layers. If we apply \reinit without adding normalization, the test accuracy \emph{drops} from about 34\% to 24\%. With the normalization layers inserted by \reinitns, it \emph{improves} to 42\%.

\paragraph{Training Longer.}
The improvement in \reinit cannot be obtained by simply training longer even with learning rate scheduling. Throughout our experiments (e.g. Tables~\ref{tab:reinit_results_with_aug} and~\ref{tab:reinit_results_no_aug}), we also train the baseline longer to have the same number of training steps in total as reinitialization methods. Despite that, reinitialization methods improve performance considerably. 

\section{Complete Figures for all Datasets}\label{appendix::full_figures}
\begin{figure*}[tbp]
    \centering
    \includegraphics[width=1.6\columnwidth]{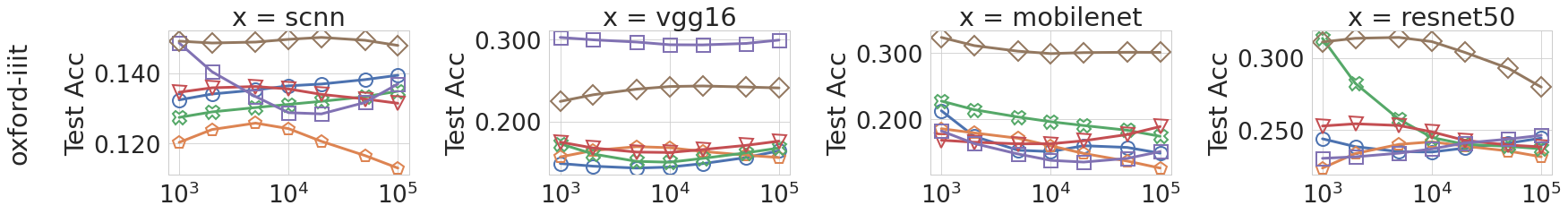}\\
    \includegraphics[width=1.6\columnwidth]{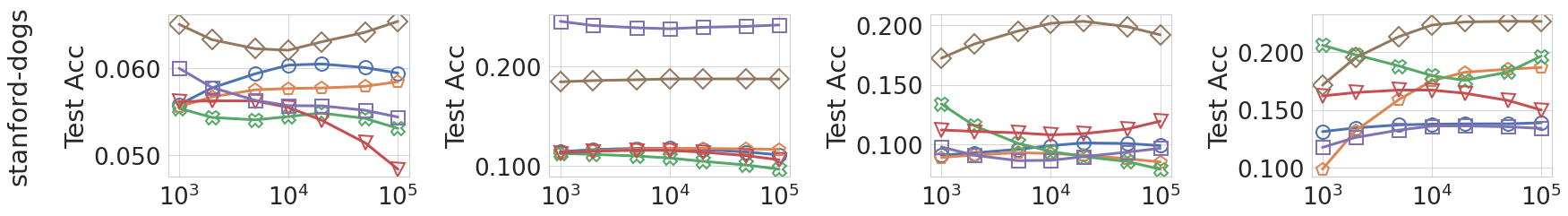}\\
    \includegraphics[width=1.6\columnwidth]{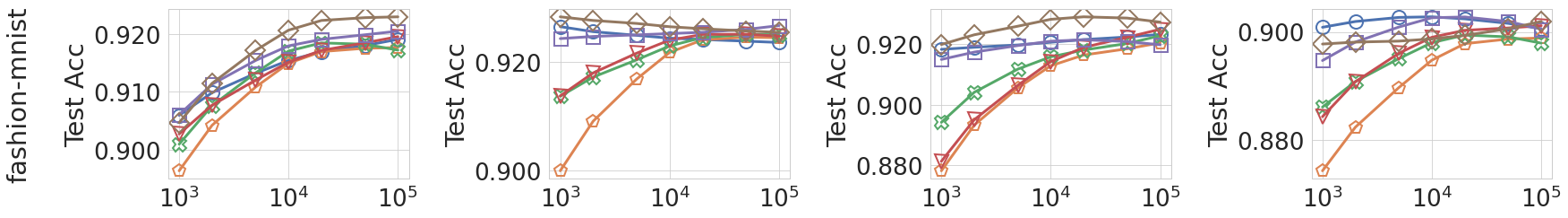}\\
    \includegraphics[width=1.6\columnwidth]{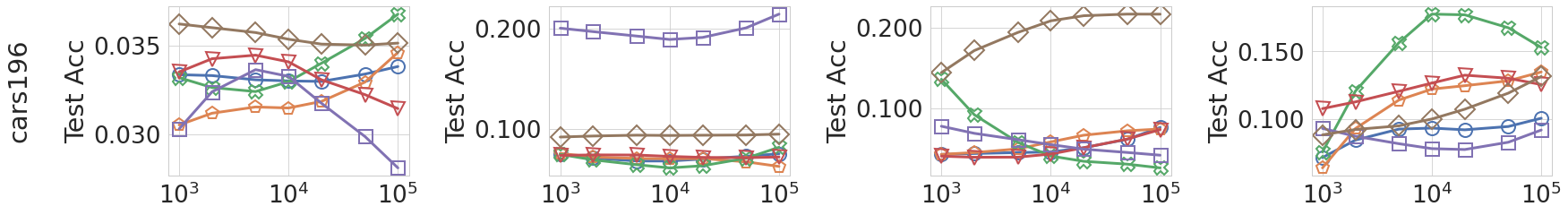}\\
    \includegraphics[width=1.6\columnwidth]{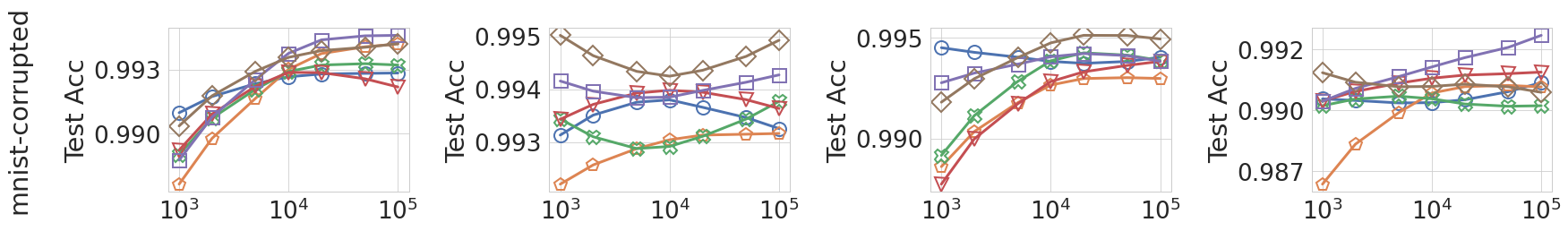}\\
    \includegraphics[width=1.6\columnwidth]{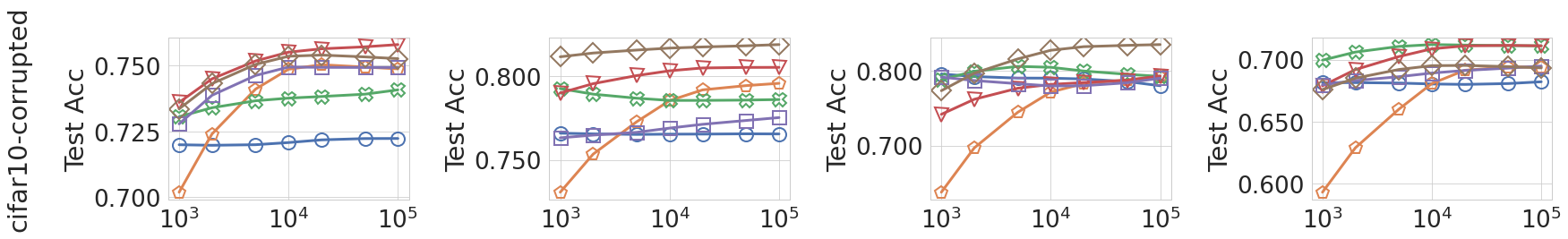}\\
    \includegraphics[width=1.6\columnwidth]{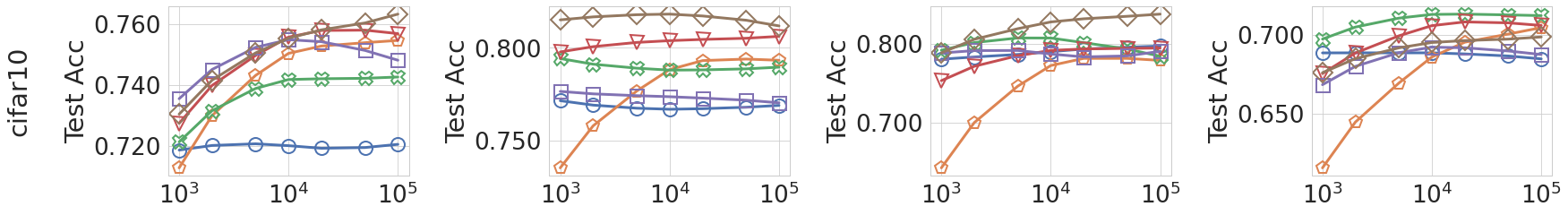}\\
    \includegraphics[width=1.6\columnwidth]{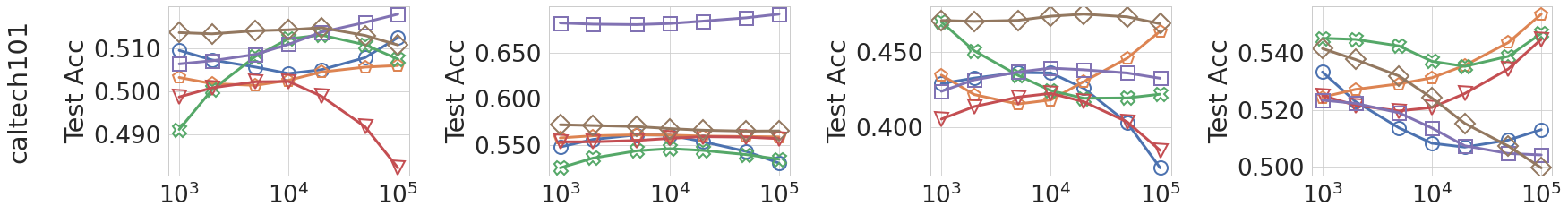}\\
    \includegraphics[width=1.6\columnwidth]{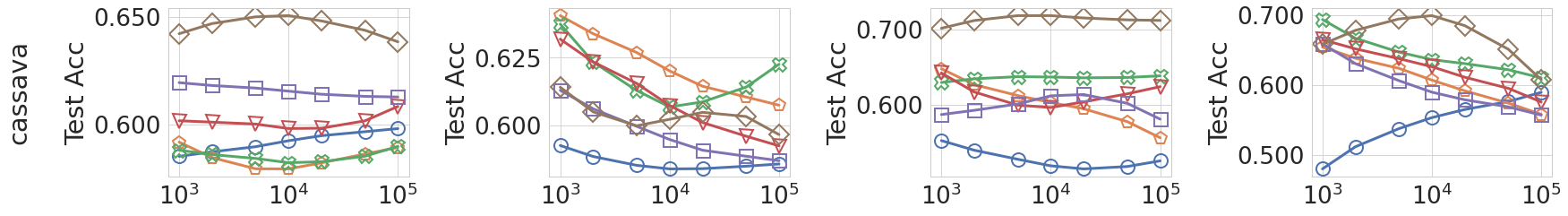}\\
    \includegraphics[width=1.6\columnwidth]{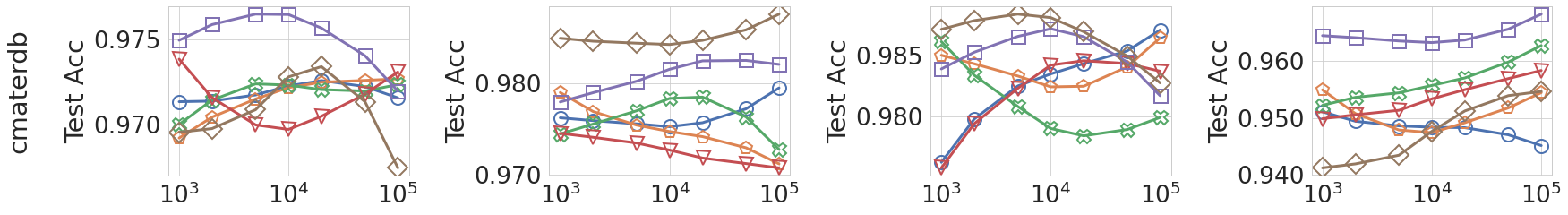}\\
    \includegraphics[width=1.6\columnwidth]{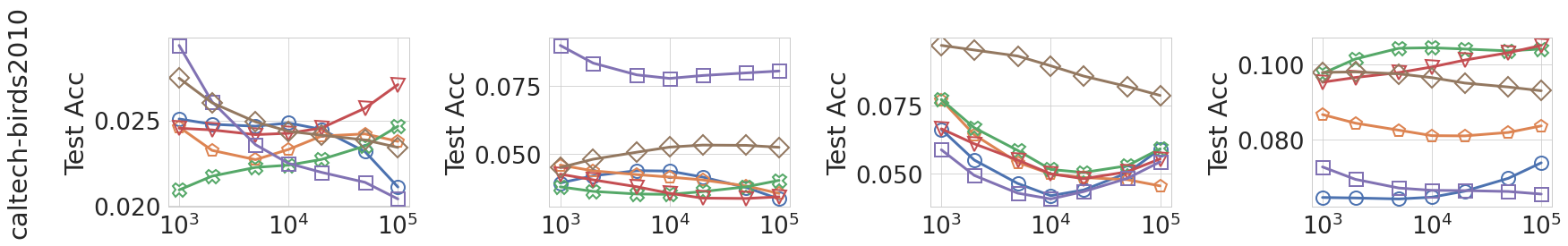}\\
    \includegraphics[width=1.6\columnwidth]{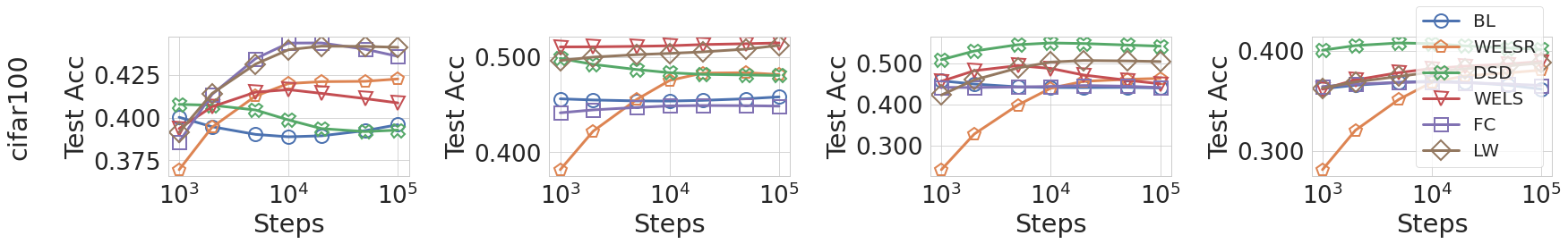}\\
    \caption{The full version of Figure \ref{fig:reinit_compute}, which contains the full set of datasets.}
    \label{fig:reinit_compute_full}
\end{figure*}

\begin{figure*}[tbp]
    \centering
    \includegraphics[width=1.6\columnwidth]{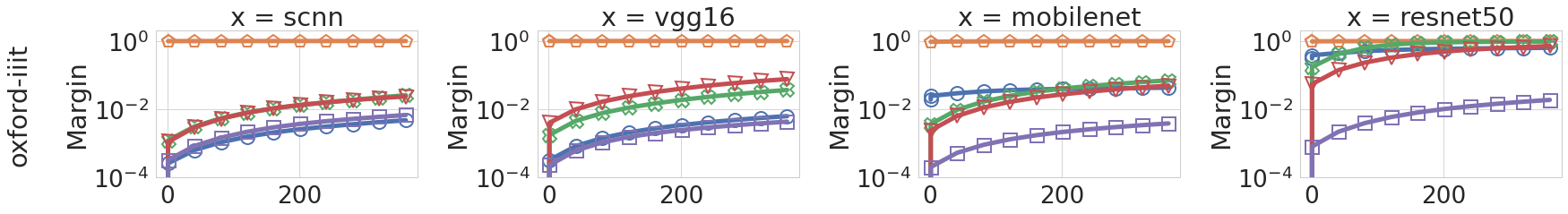}\\
    \includegraphics[width=1.6\columnwidth]{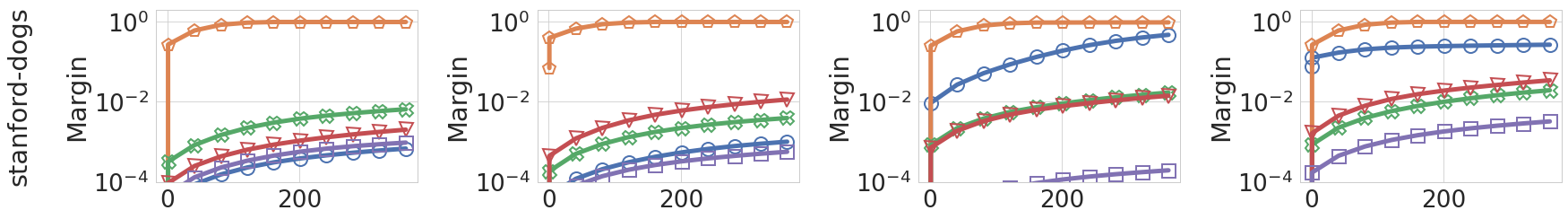}\\
    \includegraphics[width=1.6\columnwidth]{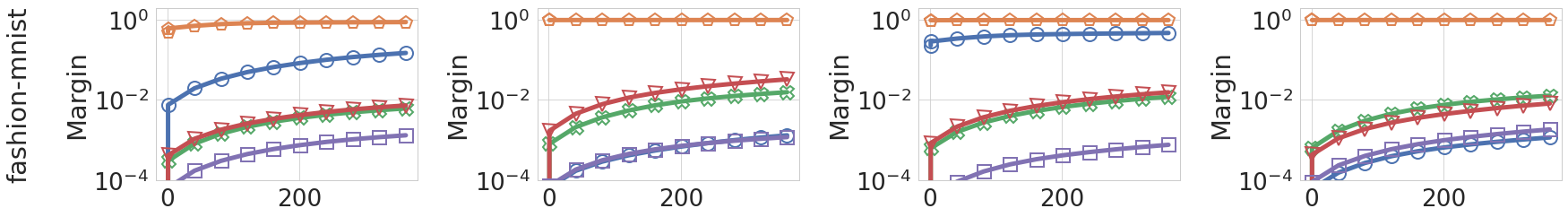}\\
    \includegraphics[width=1.6\columnwidth]{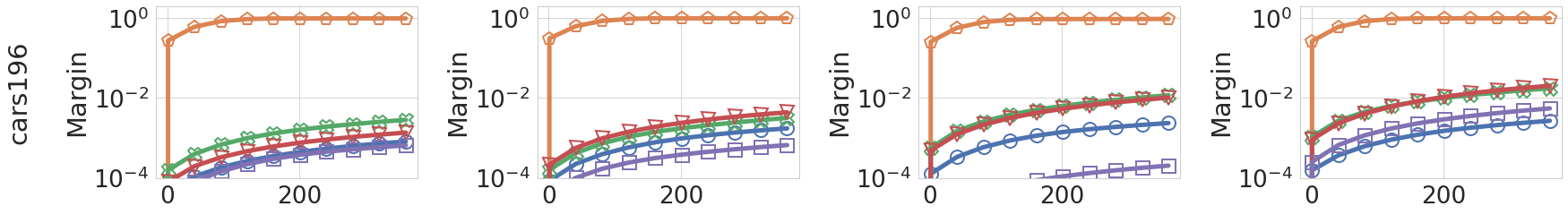}\\
    \includegraphics[width=1.6\columnwidth]{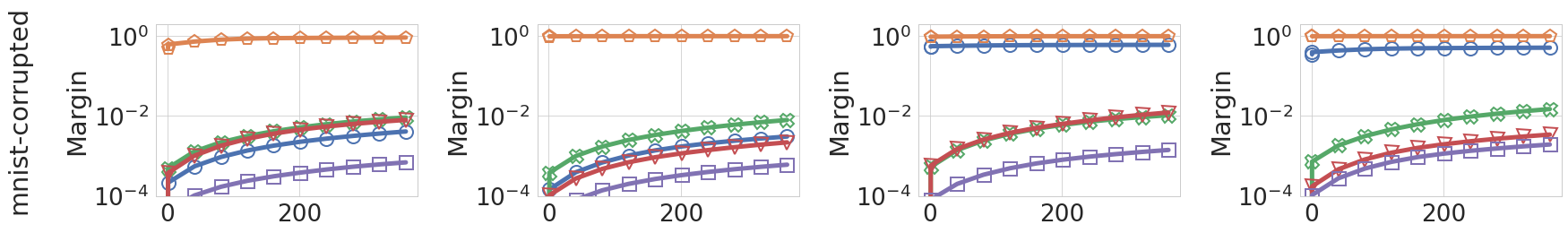}\\
    \includegraphics[width=1.6\columnwidth]{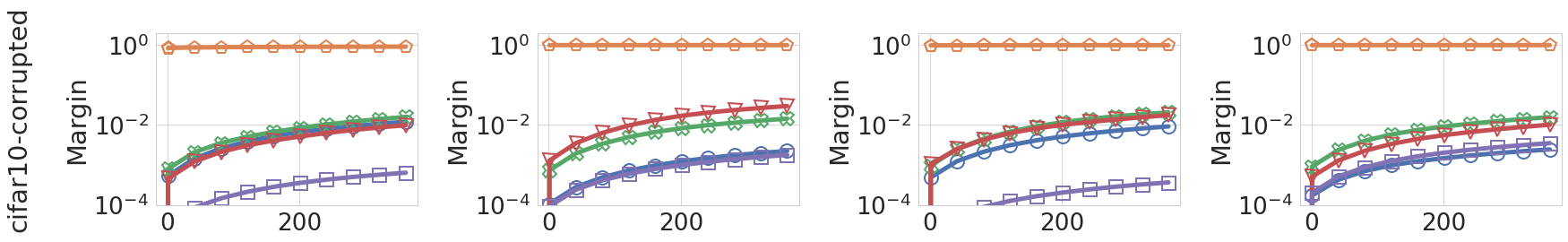}\\
    \includegraphics[width=1.6\columnwidth]{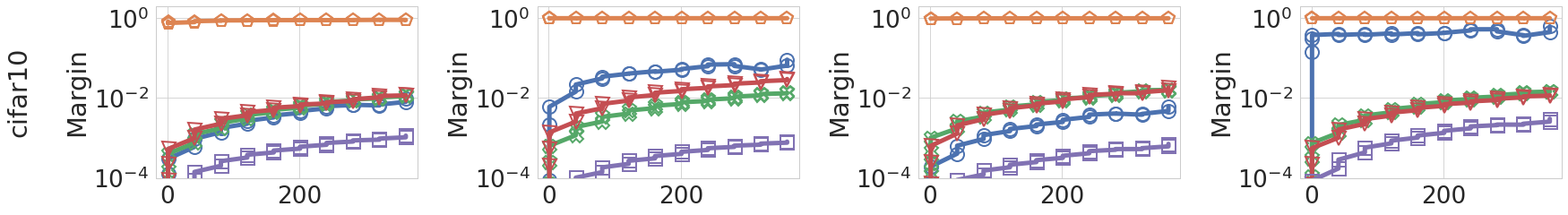}\\
    \includegraphics[width=1.6\columnwidth]{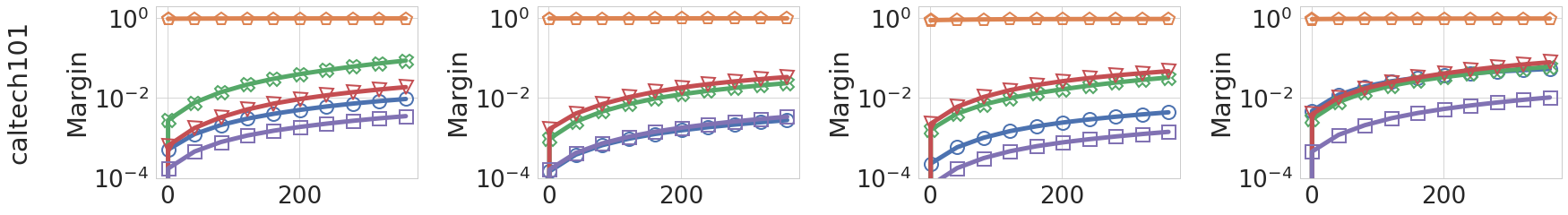}\\
    \includegraphics[width=1.6\columnwidth]{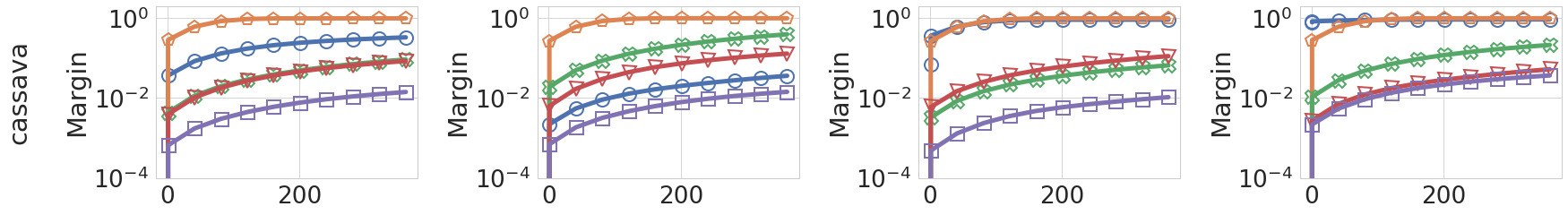}\\
    \includegraphics[width=1.6\columnwidth]{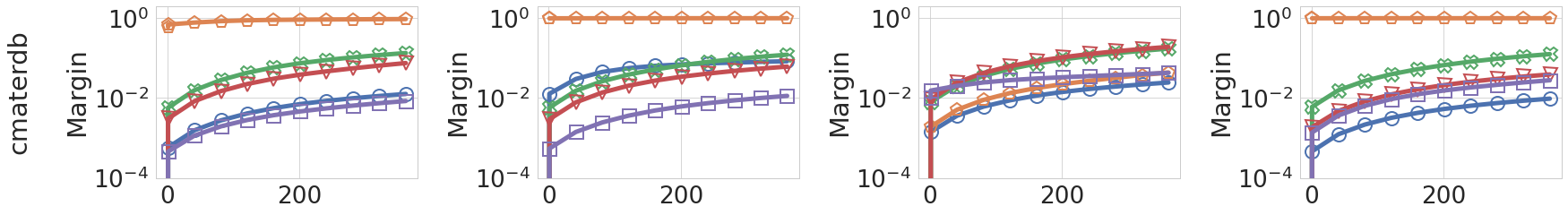}\\
    \includegraphics[width=1.6\columnwidth]{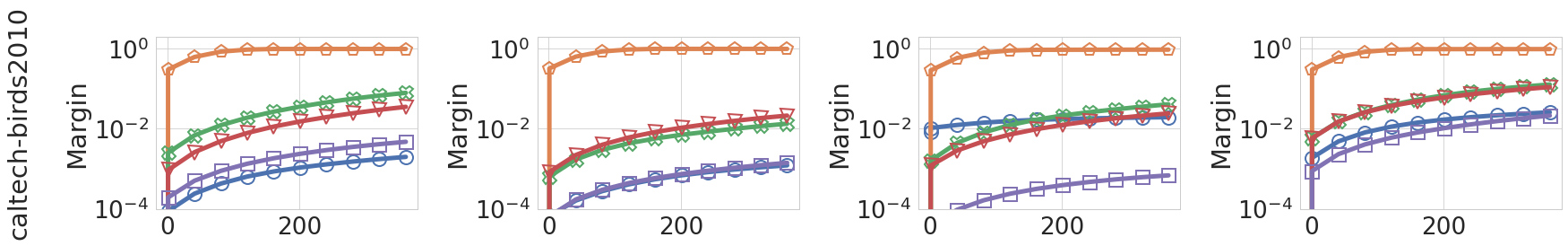}\\
    \includegraphics[width=1.6\columnwidth]{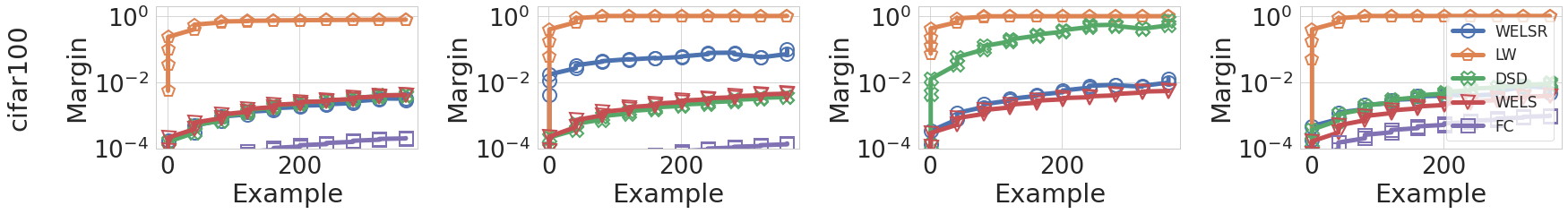}\\
    \caption{The full version of Figure \ref{fig:margins} ({\sc top}), which contains the full set of datasets. }
    \label{fig:margins)full}
\end{figure*}

\begin{figure*}[tbp]
    \centering
    \includegraphics[width=1.6\columnwidth]{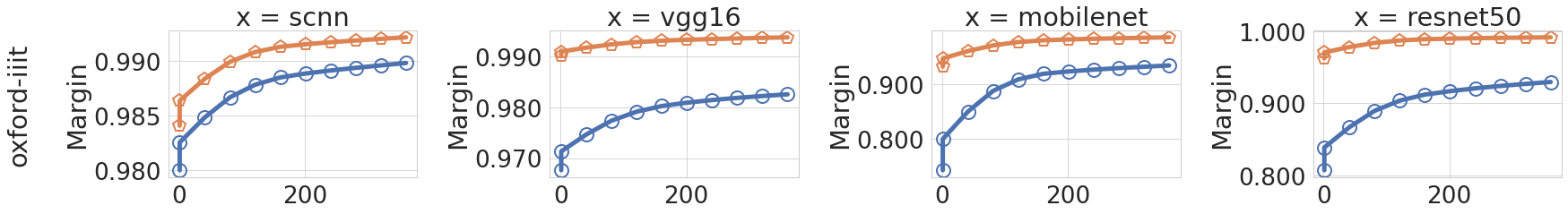}\\
    \includegraphics[width=1.6\columnwidth]{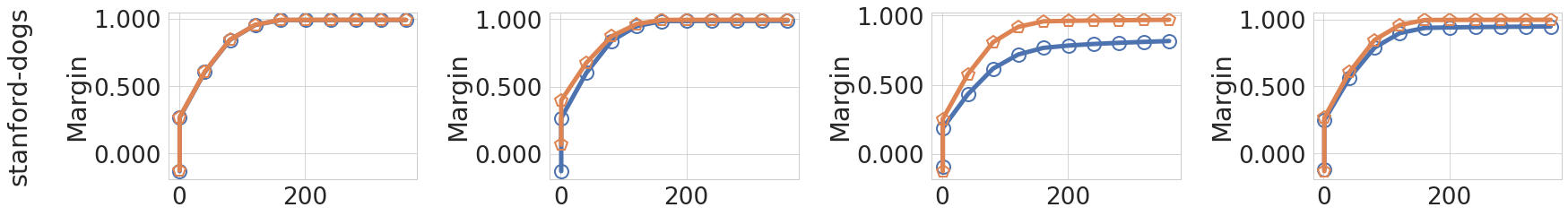}\\
    \includegraphics[width=1.6\columnwidth]{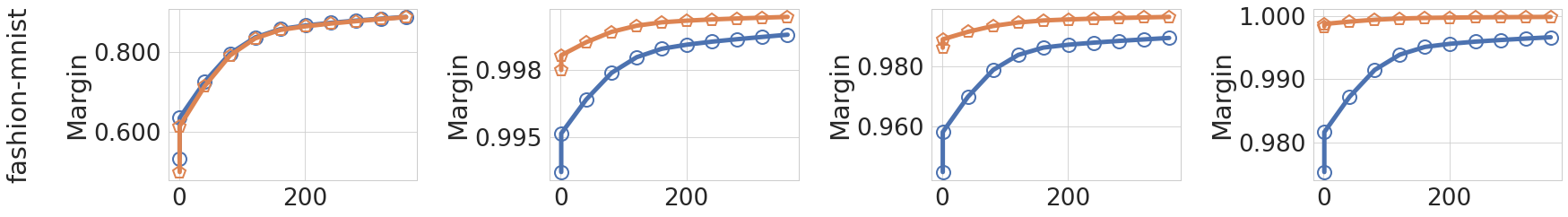}\\
    \includegraphics[width=1.6\columnwidth]{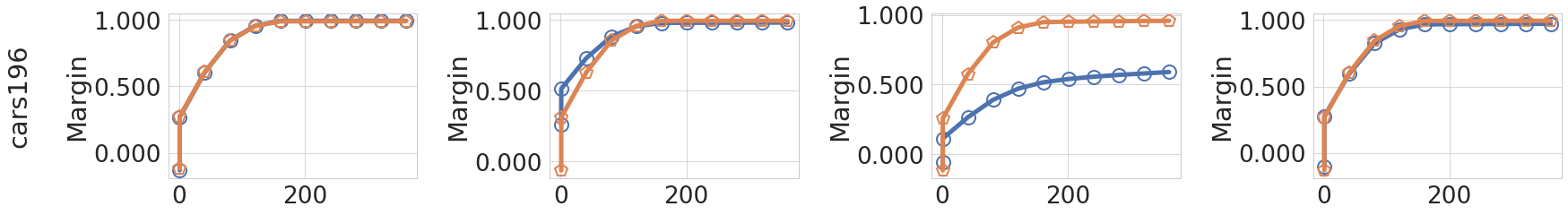}\\
    \includegraphics[width=1.6\columnwidth]{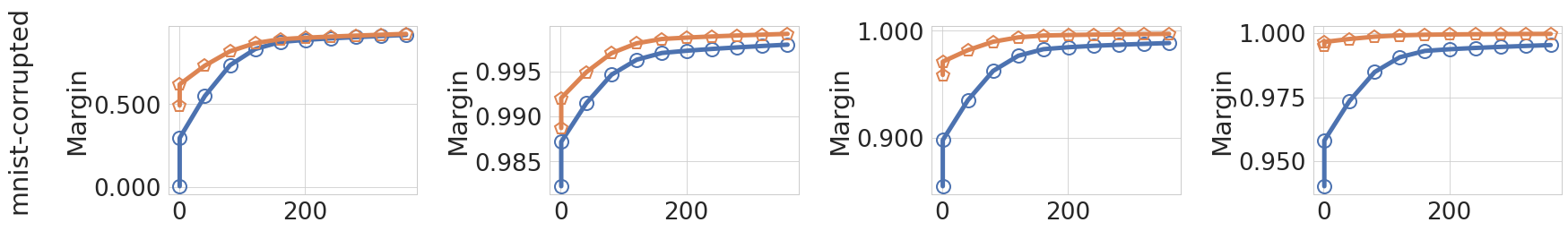}\\
    \includegraphics[width=1.6\columnwidth]{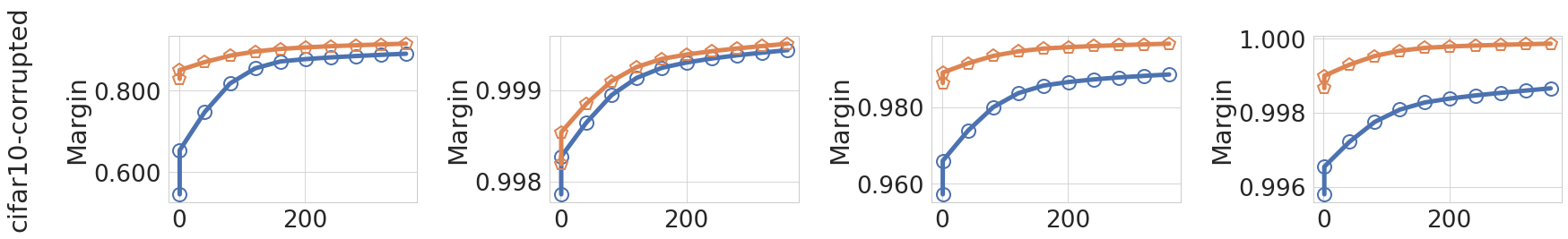}\\
    \includegraphics[width=1.6\columnwidth]{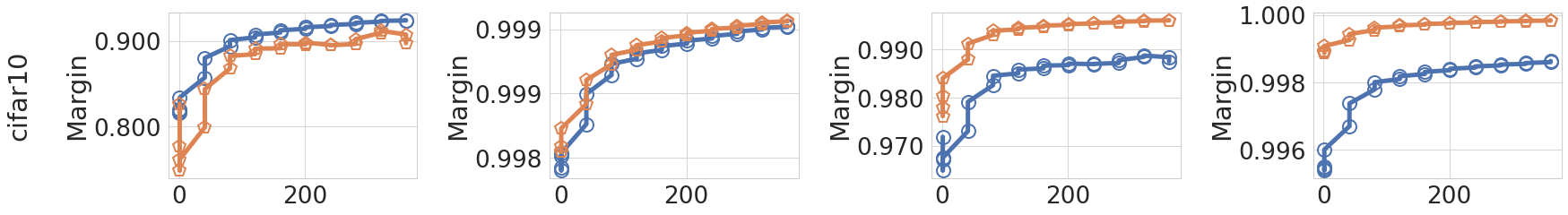}\\
    \includegraphics[width=1.6\columnwidth]{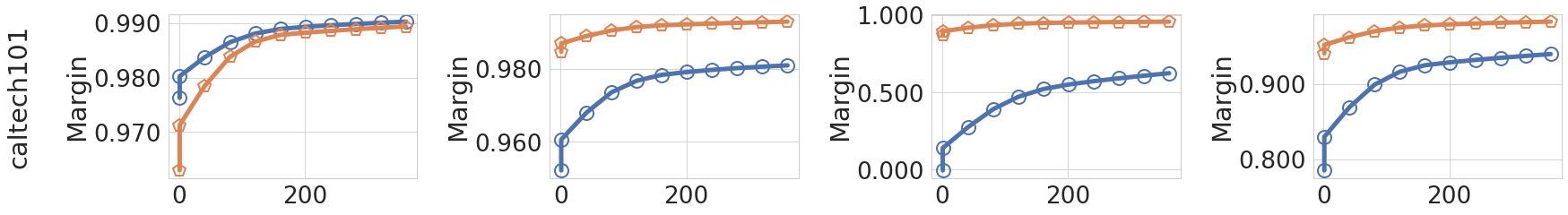}\\
    \includegraphics[width=1.6\columnwidth]{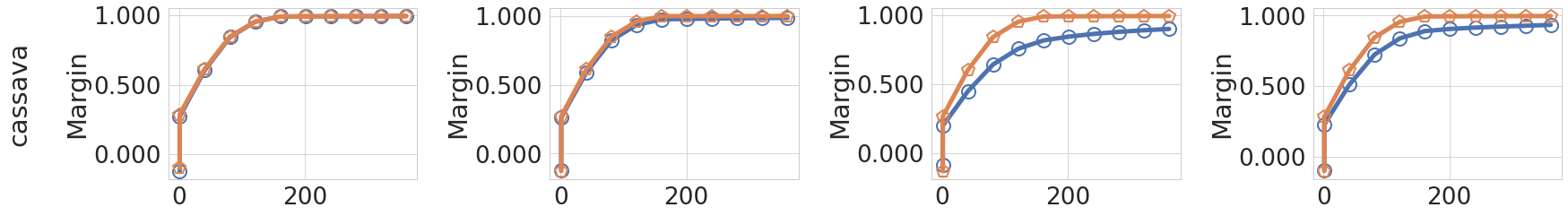}\\
    \includegraphics[width=1.6\columnwidth]{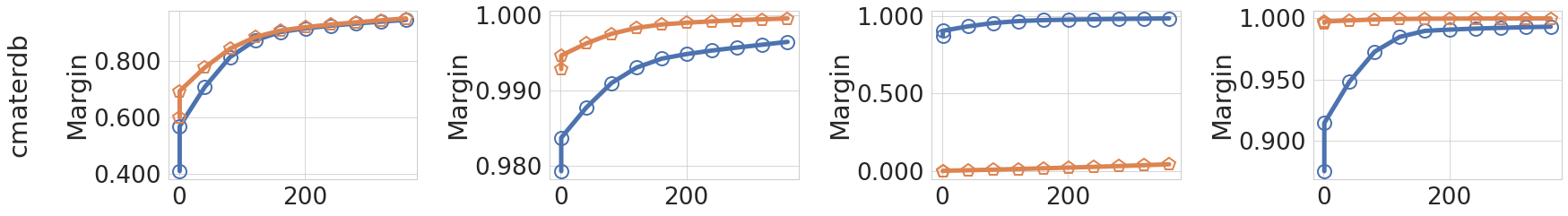}\\
    \includegraphics[width=1.6\columnwidth]{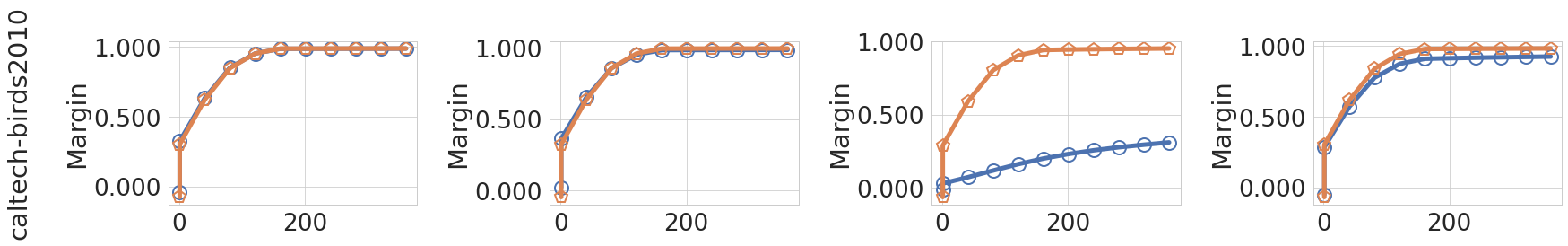}\\
    \includegraphics[width=1.6\columnwidth]{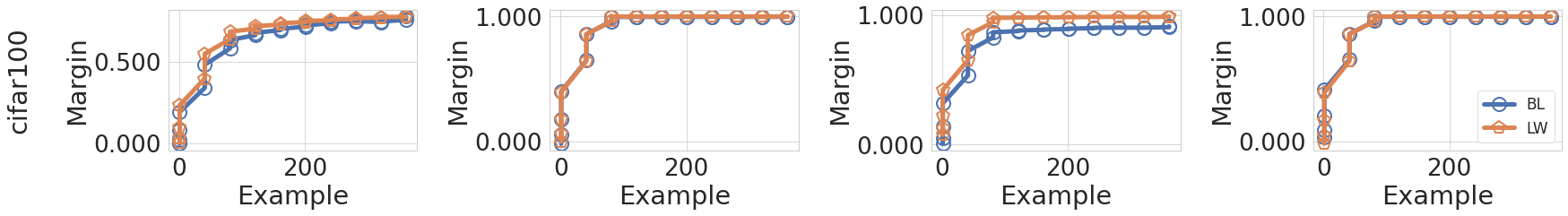}\\
    \caption{The full version of Figure \ref{fig:margins} ({\sc bottom}), which contains the full set of datasets.}
    \label{fig:margins_bl_full}
\end{figure*}

\begin{figure*}[tbp]
    \centering
    \includegraphics[width=1.6\columnwidth]{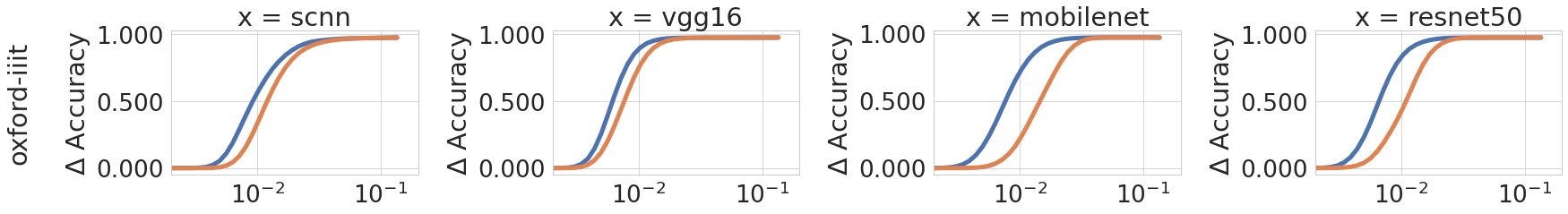}\\
    \includegraphics[width=1.6\columnwidth]{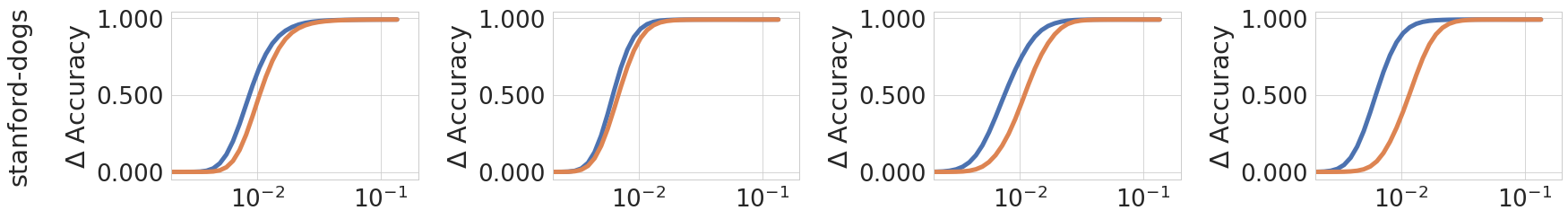}\\
    \includegraphics[width=1.6\columnwidth]{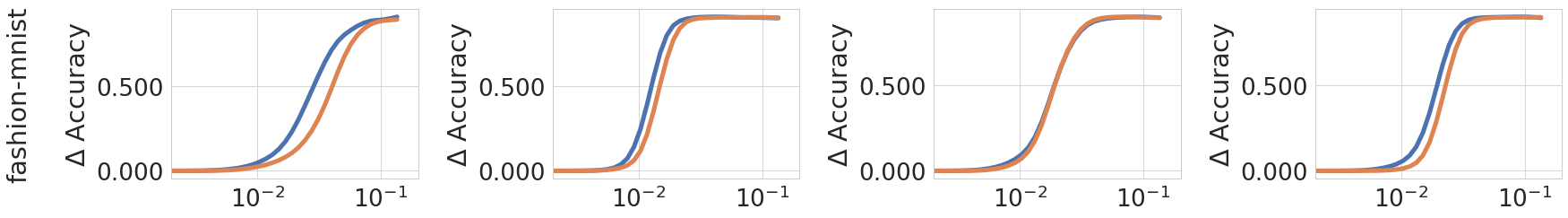}\\
    \includegraphics[width=1.6\columnwidth]{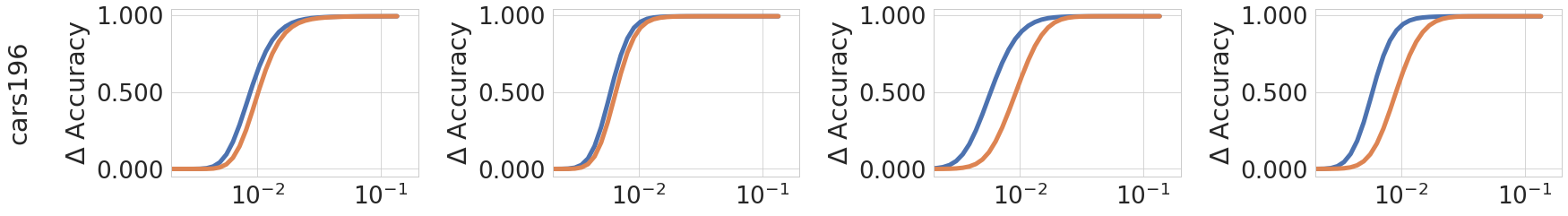}\\
    \includegraphics[width=1.6\columnwidth]{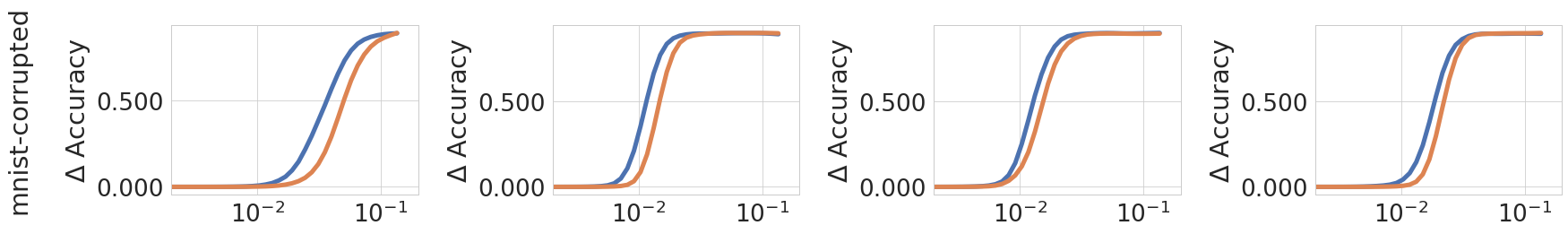}\\
    \includegraphics[width=1.6\columnwidth]{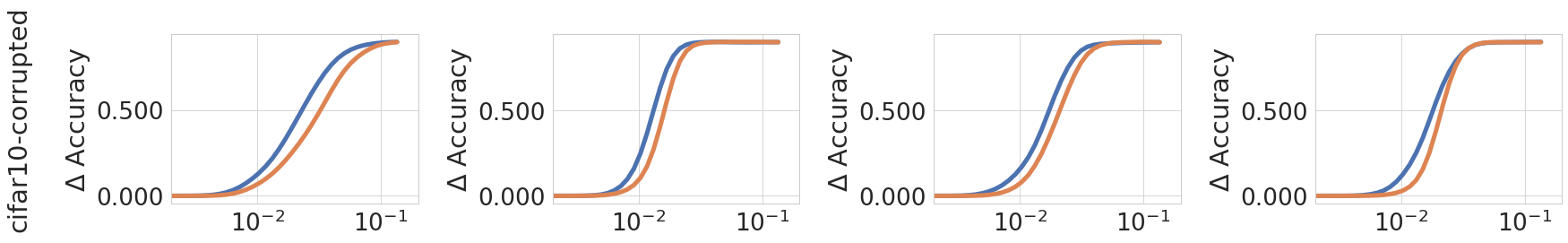}\\
    \includegraphics[width=1.6\columnwidth]{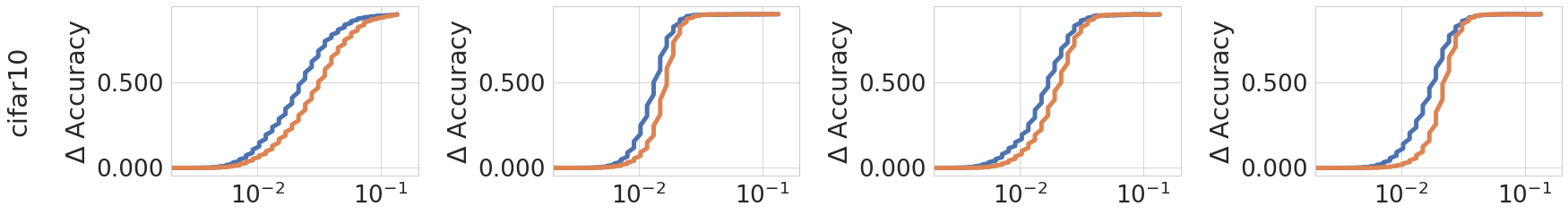}\\
    \includegraphics[width=1.6\columnwidth]{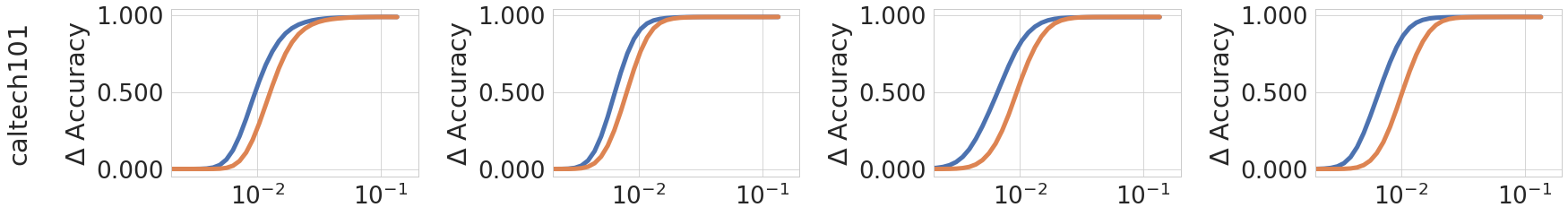}\\
    \includegraphics[width=1.6\columnwidth]{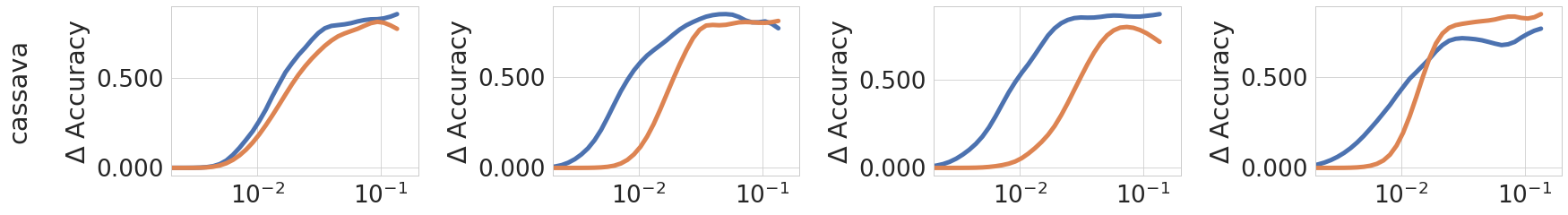}\\
    \includegraphics[width=1.6\columnwidth]{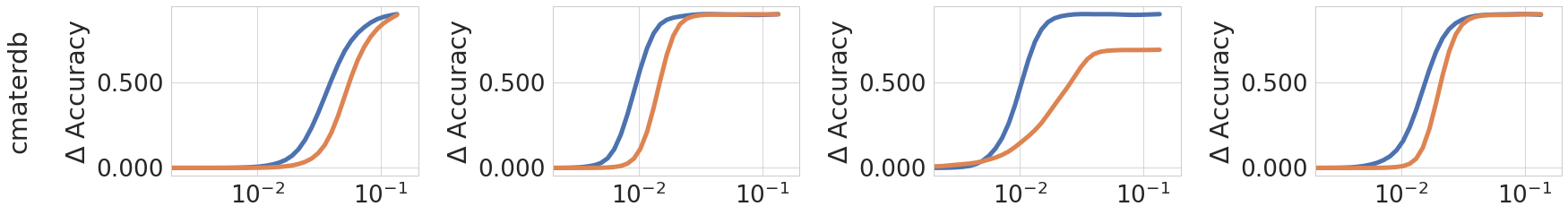}\\
    \includegraphics[width=1.6\columnwidth]{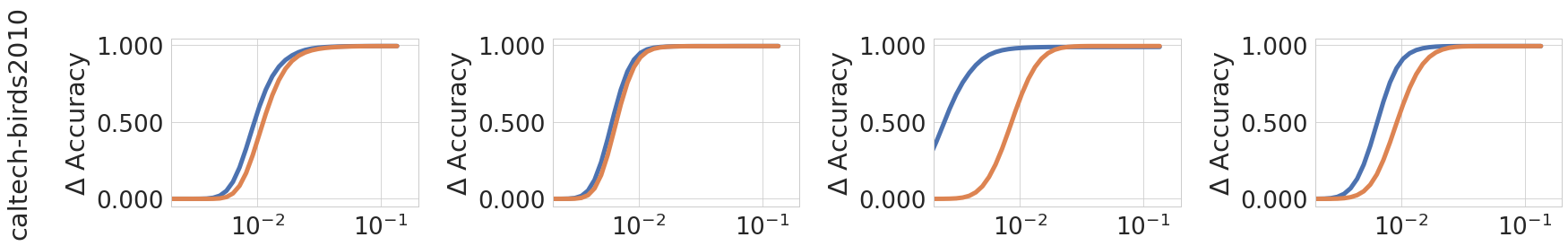}\\
    \includegraphics[width=1.6\columnwidth]{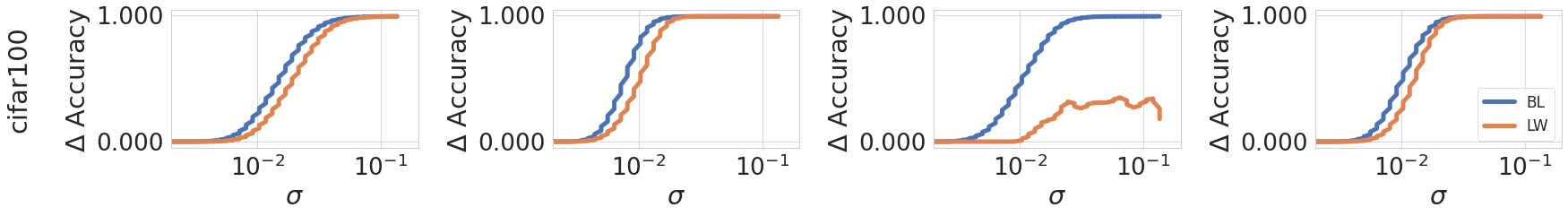}\\
    \caption{The full version of Figure \ref{fig:kl_acc}, which contains the full set of datasets.}
    \label{fig:kl_acc_full}
\end{figure*}

For completeness, we provide the full version of Figures \ref{fig:reinit_compute}, \ref{fig:margins}, and \ref{fig:kl_acc} in this section. These are provided in Figures \ref{fig:reinit_compute_full}, \ref{fig:margins)full}, \ref{fig:margins_bl_full}, and \ref{fig:kl_acc_full}. 

\clearpage

\end{document}